\documentclass[runningheads]{llncs}

\usepackage{eccv}

\usepackage{eccvabbrv}

\usepackage{graphicx}
\usepackage{booktabs}
\usepackage{algorithm}
\usepackage{placeins}
\usepackage{multirow}
\usepackage{array}
\usepackage{tabularx}

\usepackage[accsupp]{axessibility}  

\usepackage{hyperref}

\usepackage{orcidlink}

\graphicspath{{./Figs/}}

\newcommand{\para}[1]{\vspace{.05in}\noindent\textbf{#1}\quad}

\begin{document}

\title{When W4A4 Breaks Camouflaged Object Detection: Token-Group Dual-Constraint Activation Quantization} 

\titlerunning{COD-TDQ}

\author{Tianqi Li\inst{1}\orcidlink{0009-0007-5978-0389} \and
Wenyu Fang\inst{1}\orcidlink{0009-0005-8847-3017} \and
Xin He\inst{2}\orcidlink{0009-0004-2139-6590} \and
Xue Geng\inst{3}\orcidlink{0000-0002-2594-9648} \and
Xu Cheng\inst{2}\orcidlink{0000-0002-4724-5748} \and 
Yun Liu\inst{1,4,5}\thanks{Corresponding author: Yun Liu (liuyun@nankai.edu.cn)}\orcidlink{0000-0001-6143-0264}}

\authorrunning{Li et al.}


\institute{VCIP, College of Computer Science, Nankai University \and
School of Computer Science and Engineering, Tianjin University of Technology \and
Institute for Infocomm Research, A*STAR \and
Academy for Advanced Interdisciplinary Studies, Nankai University \and
Nankai International Advanced Research Institute, Shenzhen Futian}

\maketitle

\begin{abstract}
  Camouflaged object detection (COD) segments objects that intentionally blend with the background, so predictions depend on subtle texture and boundary cues. COD is often needed under tight on-device memory and latency budgets, making low-bit inference highly desirable. However, COD is unusually hard to quantize aggressively. We study post-training W4A4 quantization of Transformer-based COD and find a task-specific cliff: heavy-tailed background tokens dominate a shared activation range, inflating the step size and pushing weak-but-structured boundary cues into the zero bin. This exposes a token-local bottleneck---remove cross-token \emph{range domination} and bound the \emph{zero-bin mass} under 4-bit activations. To address this, we introduce \textbf{COD-TDQ}, a \textbf{COD}-aware \textbf{T}oken-group \textbf{D}ual-constraint activation \textbf{Q}uantization method. COD-TDQ addresses this token-local bottleneck with two coupled steps: \textbf{D}irect-\textbf{S}um \textbf{T}oken-\textbf{G}roup (\textbf{DSTG}) assigns \emph{token-group} scales to suppress cross-token range domination, and \textbf{D}ual-\textbf{C}onstraint \textbf{R}ange \textbf{P}rojection (\textbf{DCRP}) projects each token-group clip range to keep the step-to-dispersion ratio and the zero-bin mass bounded. Across four COD benchmarks and two baseline models (CFRN and ESCNet), COD-TDQ consistently achieves an $S_\alpha$ score more than 0.12 higher than that of the state-of-the-art quantization method without retraining. The code is available at \href{https://github.com/MCG-NKU/nku-model-compre}{https://github.com/MCG-NKU/nku-model-compre}.
\end{abstract}

\section{Introduction}
\label{sec:intro}
Camouflaged object detection (COD) is typically evaluated as binary mask prediction for objects that intentionally blend into the background.
COD models must rely on subtle texture and boundary cues, so useful evidence often appears as small-amplitude yet structured responses~\cite{backgroundIOC}.
As Transformer~\cite{att,ViT,liu2024vision} encoders and multi-stage designs become common~\cite{guo2023visual,CamoFormer,LersGAN,ICIC,zhou2025sam2,chen2025enhancing,ariff2026evaluating}, COD models have witnessed significant development~\cite{OVCOD,Du2025ShiftTL,Zero-Shot,zhou2025rethinking,samcod,MMSA,ren2025multi,zhao2026open,ji2024segment,li2024novel,wu2026gapnet}. However, with this growth, COD models are becoming more complex, increasing deployment memory and latency costs~\cite{CamoDiffusion, Controllable-LPMoE, Liu2025ImprovingSF, WPAQTP, Camouflage}.
At the same time, COD is increasingly expected to operate under strict memory and latency constraints~\cite{WU2025110771,REN2025113056,ESN,gao2026catp}, especially for applications on mobile and edge devices. In this context, it becomes desirable to reduce the computational and storage costs of models during deployment. One practical solution is to apply post-training quantization (PTQ), which can effectively reduce memory usage and activation cost without requiring retraining~\cite{Banner2018PostT4}.

In practice, INT8 PTQ is often the default choice~\cite{frantar2022gptq, xu2026parameter}. Nevertheless, when deployment budgets are extremely limited, the gains from INT8 can be modest, and further reductions require moving to ultra-low-bit regimes. This motivates exploring ultra-low-bit PTQ for COD while maintaining accuracy. In particular, 4-bit weights and 4-bit activations (W4A4) reduce deployment cost under a standard protocol~\cite{stdprotocolLRPQViT, Banner2018PostT4} without changing the training or inference pipeline.
However, we find that naive W4A4 on Transformer-based COD can collapse rather than degrade smoothly.
This raises a key question: \textit{what makes COD fragile at 4-bit activations, and how can we make W4A4 reliable without retraining or hardware-specific assumptions?}

Transformer PTQ has advanced rapidly, but most pipelines still fall into a few recurring designs:
(i) block-wise reconstruction and rounding to match FP32 outputs on calibration data~\cite{yuan2022ptq4vit, li2023repqvit, wu2025fimaq},
(ii) range re-parameterization and smoothing to mitigate heavy-tailed activations~\cite{xiao2023smoothquant},
and (iii) finer granularity via grouping or adaptive scales~\cite{moon2024igqvit}.
These strategies are largely layer or block-centric and optimize average fidelity.
In COD at W4A4, the dominant failure is token-local: token-wise activation heterogeneity lets background tokens dominate the range and increase the zero-bin mass, which layer-wise fitting or reconstruction does not explicitly bound.
\begin{figure}[t]
\centering
\includegraphics[width=\linewidth]{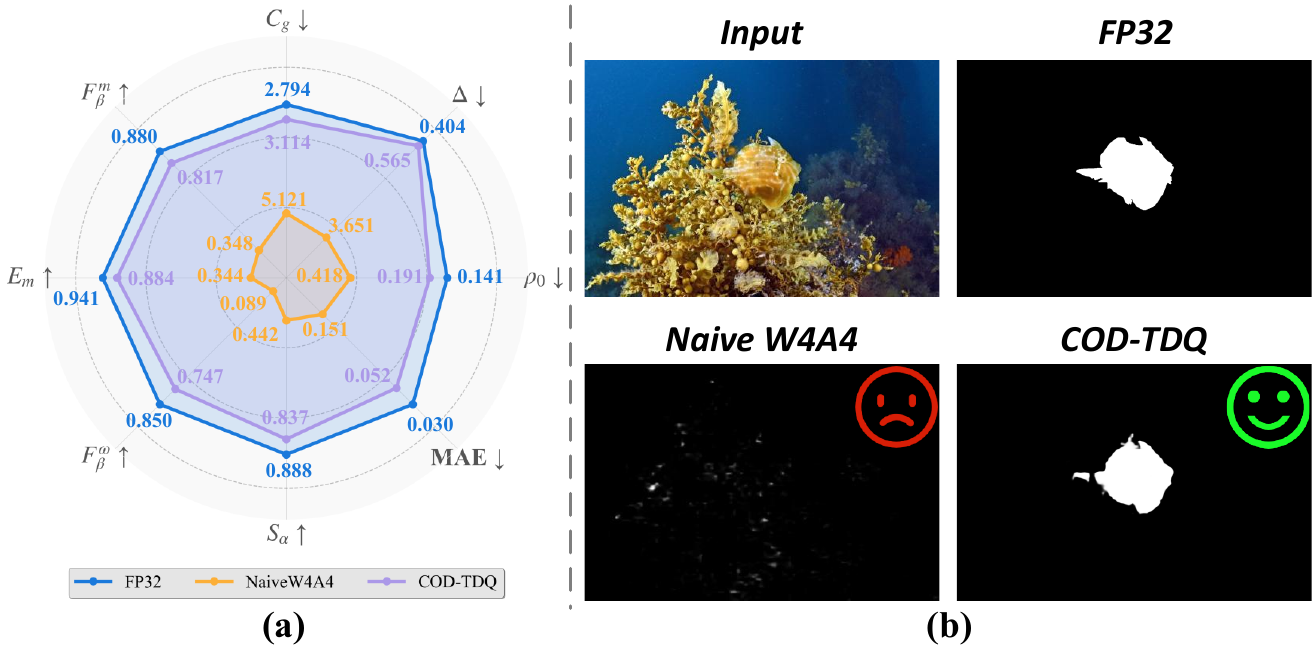}
\caption{\textbf{COD-specific W4A4 failure.} Naive W4A4 inflates a shared clipping range, producing a coarse step size and high zero-bin mass that erases weak boundary evidence. The inset summarizes representative diagnostics \((c_g,\Delta,\rho_0)\) and the associated \(S_\alpha\) collapse/recovery on CFRN/NC4K (rounded to three decimals).}
\label{fig:failure_mechanism}
\end{figure}

We trace the W4A4 cliff to a coupled collapse mechanism of \textbf{range domination} and \textbf{zero-bin mass}  (\cref{fig:failure_mechanism}).
Background tokens with heavy-tailed activation spikes can dominate a shared clipping range, inflating the step size $\Delta$ and coarsening quantization for most tokens (\cref{fig:failure_mechanism}).
Under rounding-to-nearest, weak yet structured boundary responses fall into the zero bin, yielding a high $\rho_0=\mathbb{P}(|x_{\text{boundary}}|\le \Delta/2)$. Once zeroed, subsequent attention mixing cannot recover the missing signed evidence.
The collapse is activation-dominated (W4A8 stays close to FP32 while W4A4 fails on CFRN~\cite{song2025cfrn}, as shown in~\cref{tab:motivation}), suggesting that COD needs token-local range control that explicitly bounds both the resolution ratio $\eta=\Delta/\sigma$ and the zero-bin mass $\rho_0$ (\cref{fig:dcrp_vis}).

\begin{table}[t]
\centering
\caption{\textbf{Motivating observation} on NC4K (CFRN backbone).}
\label{tab:motivation}
\resizebox{0.75\textwidth}{!}{ 
\begin{tabular}{lccccc}
\toprule
Setting & $S_\alpha\uparrow$ & $F^{\omega}_{\beta}\uparrow$ & $E_m\uparrow$ & $F^m_\beta\uparrow$ & MAE$\downarrow$ \\
\midrule
FP32 (baseline) & \textbf{.888} & \textbf{.850} & \textbf{.941} & \textbf{.880} & \textbf{.030} \\
W8A8 (naive)    & \underline{.887} & \underline{.850} & \underline{.940} & \underline{.879} & \underline{.030} \\
W4A8 (naive)    & .882 & .841 & .936 & .872 & .032 \\
W4A4 (naive)    & .443 & .089 & .344 & .348 & .151 \\
\bottomrule
\end{tabular}
}
\end{table}

Based on diagnosis, we propose \textbf{COD-TDQ}, a COD-aware dynamic activation quantization framework for W4A4 Transformer-based COD.
COD-TDQ remains purely PTQ (no retraining) and is hardware-agnostic.
It combines Direct-Sum Token-Group (\textbf{DSTG}) to assign token-group activation scales and remove cross-token \textbf{range domination}, and Dual-Constraint Range Projection (\textbf{DCRP}) to project each token-group range so that the step ratio $\eta$ and the \textbf{zero-bin mass} $\rho_0$ stay in a stable regime.
Across four COD benchmarks and two Transformer COD models (CFRN and ESCNet~\cite{ye2025escnet}), we perform comprehensive and extensive evaluations of COD-TDQ. The results consistently show that COD-TDQ significantly surpasses representative PTQ baselines under the same W4A4 quantization protocol, delivering over 0.12–0.14 improvements in $S_\alpha$ (CFRN) compared with the state-of-the-art quantization approach without retraining.

Our contributions can be summarized as follows:

\begin{itemize}
    \item We provide a mechanism-driven diagnosis of COD-specific W4A4 collapse, centered on \textbf{range domination} and \textbf{zero-bin mass} collapse, with diagnostic metrics of $\Delta$, $\eta$ and $\rho_0$, respectively.
    \item We propose \textbf{COD-TDQ}, which stabilizes W4A4 activation ranges via (i) Direct-Sum Token-Group scaling (DSTG) to suppress \textbf{range domination}, and (ii) Dual-Constraint Range Projection (DCRP), which enforces \textbf{step-to-dispersion} and \textbf{zero-bin-mass} constraints to bound $\eta$ and $\rho_0$.
    \item We establish a unified W4A4 PTQ benchmark for COD across four datasets and provide diagnostic metrics (i.e., $\Delta$, $\eta$, $\rho_0$) that explain baseline failures, facilitating future research on COD quantization.
\end{itemize}

\section{Related Work}
\label{sec:related_work}
\noindent\textbf{PTQ for Transformers and Segmentation Models.}\quad
We focus on W4A4 PTQ for Transformer-based COD and relate our work to recent advances in Transformer PTQ.
PTQ aims to convert pretrained networks to low precision without retraining, typically combining weight quantization with activation range selection using a calibration set and local reconstruction \cite{li2024branch,zhang2025selectq}.
For vision transformers(ViT)~\cite{ViT}, PTQ4ViT~\cite{yuan2022ptq4vit} exemplifies block-level calibration and reconstruction to make projections quantizable.
Subsequent methods such as RepQ-ViT~\cite{li2023repqvit} and FIMA-Q~\cite{wu2025fimaq} further reduce the quantization-induced representation drift by improving feature fidelity.
Optimization-based rounding (AdaRound~\cite{nagel2020adaround}), block reconstruction with scheduling (BRECQ~\cite{li2021brecq}), and calibration regularization (QDrop~\cite{wei2022qdrop}) which minimize reconstruction error around the calibration distribution.
PQ-SAM~\cite{liu2024pqsam} and PTQ4SAM~\cite{lv2024ptq4sam} study PTQ strategies tailored to the Segment Anything Model (SAM), emphasizing sensitive pathways in attention and normalization and the interaction between prompt encoders and mask decoders.
But the COD setting differs: informative signals often appear as small-amplitude, structured boundary responses embedded in background-driven heavy tails.
This shifts the bottleneck from global fidelity to preserving weak token-local cues under aggressive activation quantization.

\para{Token Sensitivity and Activation Heterogeneity.}
Ultra-low-bit quantization is particularly sensitive to heavy-tailed activations and heterogeneous statistics.
ORQ-ViT~\cite{ning2025orqvit} explicitly handles outliers to prevent a small number of extreme values from dominating the clipping range, while NoisyQuant~\cite{yang2023noisyquant} introduces noise and perturbation modeling to better match non-Gaussian activation behaviors.
SmoothQuant~\cite{xiao2023smoothquant} re-parameterizes activations and weights to ease activation quantization by shifting difficulty to weights, and AHCPTQ~\cite{zhang2025ahcptq} treats activation heterogeneity as a first-class issue in calibration.
The post-GELU token-based dynamic bit-width assignment~\cite{kim2026tokenbitwidth} is representative in exploiting token statistics to decide where additional precision is needed. Such methods provide useful insight that token distributions are highly non-uniform, yet their main lever is changing the bit-width.
IGQ-ViT~\cite{moon2024igqvit} and ADFQ-ViT~\cite{JIANG2025107289} explore adaptive grouping strategies to refine quantization granularity beyond a single global scale.
Despite their different mechanisms, most of these methods still operate with shared ranges at the layer and block level and evaluate primarily on classification. Under COD at W4A4, the critical failures are token-local: weak boundary evidence can collapse into the zero bin, which is not directly implied by outlier suppression or average reconstruction fidelity. 
In our setting, the quantization budget is fixed (W4A4) under a standard quantization protocol, so the dominant requirement becomes stabilizing \textit{4-bit activations} without relying on dynamic bit-width execution. Bit allocation alone does not prevent token-local inflation or guarantee that minority boundaries avoid zeroization.

\para{Architecture-aware error control.}
Cross-domain PTQ designs highlight the importance of structural constraints: ARCQuant~\cite{ma2026arcquant} tailors quantization to residual and attention structures on NVFP4~\cite{abecassis2025pretraining}, and QuaRTZ~\cite{kim2025quartz} emphasizes controlling error accumulation and sparsity patterns.
While these ideas motivate architecture-aware quantization, they are not formulated around dense prediction failures driven by token-local activation range selection.
In COD under W4A4, cross-token range domination inflates the quantization step for most tokens, and zero-bin mass collapse erases weak-but-structured boundary cues.
COD-TDQ addresses this gap with \textit{token-local, channel-group-wise} activation quantization (DSTG) and a \textit{dual-constraint} range projection (DCRP) that bounds both the step-to-dispersion ratio and the zero-bin mass.

\section{W4A4 Fragility Diagnosis for COD}
\label{sec:cod_failure}
This section provides a diagnosis for why Transformer-based COD is unusually fragile under post-training W4A4. The collapse forms a coupled loop: background-dominated activation statistics inflate a shared range and step size, which increases the zero-bin mass and erases weak boundary cues that COD relies on.

\subsection{Preliminaries}
\label{sec:failure_prelim}
The failure mechanism is easiest to expose under a conventional tensor-wise activation quantizer that shares a single range across all tokens. This subsection defines the diagnostic variables used throughout the analysis.

Consider an activation tensor $X\in\mathbb{R}^{T\times C}$ from a single layer and a single input sample, where $T$ indexes tokens and $C$ indexes channels.
Let $x_{c}$ denote an element.
A symmetric $a$-bit uniform quantizer selects a tensor-wise clip radius $c>0$ and uses the integer grid $q_{\min}=-2^{a-1}$ and $q_{\max}=2^{a-1}-1$.
For W4A4 activations, $a=4$ so $q_{\max}=7$.
The corresponding step size is $\Delta=c/q_{\max}$.
Under rounding-to-nearest, the dequantized value satisfies $\hat{x}_{c}=0$ if and only if $|x_{c}|\le \Delta/2$.
Four scalar diagnostics summarize the key effects.

Global clip factor $c_g$.
To compare clip ranges across layers with different scales, the clip radius is normalized by the tensor dispersion.
Let $\sigma_g=\operatorname{Std}(X)+\varepsilon$ denote the standard deviation over all $TC$ elements, where $\varepsilon>0$ is a small constant.
The normalized global clip factor is
\begin{equation}
c_g \triangleq \frac{c}{\sigma_g},\quad
\Delta \triangleq \frac{c}{q_{\max}},\quad
\eta \triangleq \frac{\Delta}{\sigma_{\mathcal{A}}}.
\label{eq:cg_Delta_eta_oneline}
\end{equation}
Larger $c_g$ indicates that a wider range is allocated relative to the typical tensor.

Step size $\Delta$.
The step size is the quantization resolution within the unclipped region.
For tensor-wise symmetric quantization, it is defined above, where $q_{\max}$ is the maximum representable quantized magnitude.

Step-to-dispersion ratio $\eta$.
A small absolute step can still be coarse for weak cues.
For a selected activation set $\mathcal{A}$ with dispersion $\sigma_{\mathcal{A}}=\operatorname{Std}(\mathcal{A})+\varepsilon$, the resolution ratio is defined above.
$\mathcal{A}$ is instantiated as boundary-heavy activations.

Zero-bin mass $\rho_0$.
For a set of activations $\mathcal{A}$, the zero-bin mass is the fraction of elements that quantize to zero, where the probability view refers to the empirical distribution of the selected activations:
\begin{equation}
\rho_0 \triangleq \frac{1}{|\mathcal{A}|}\sum_{x\in \mathcal{A}}\mathbf{1}(\hat{x}=0)
=
\mathbb{P}\!\left(|x|\le \frac{\Delta}{2}\right).
\label{eq:rho0_def}
\end{equation}

\begin{figure*}[!t]
\centering
\includegraphics[width=0.9\textwidth]{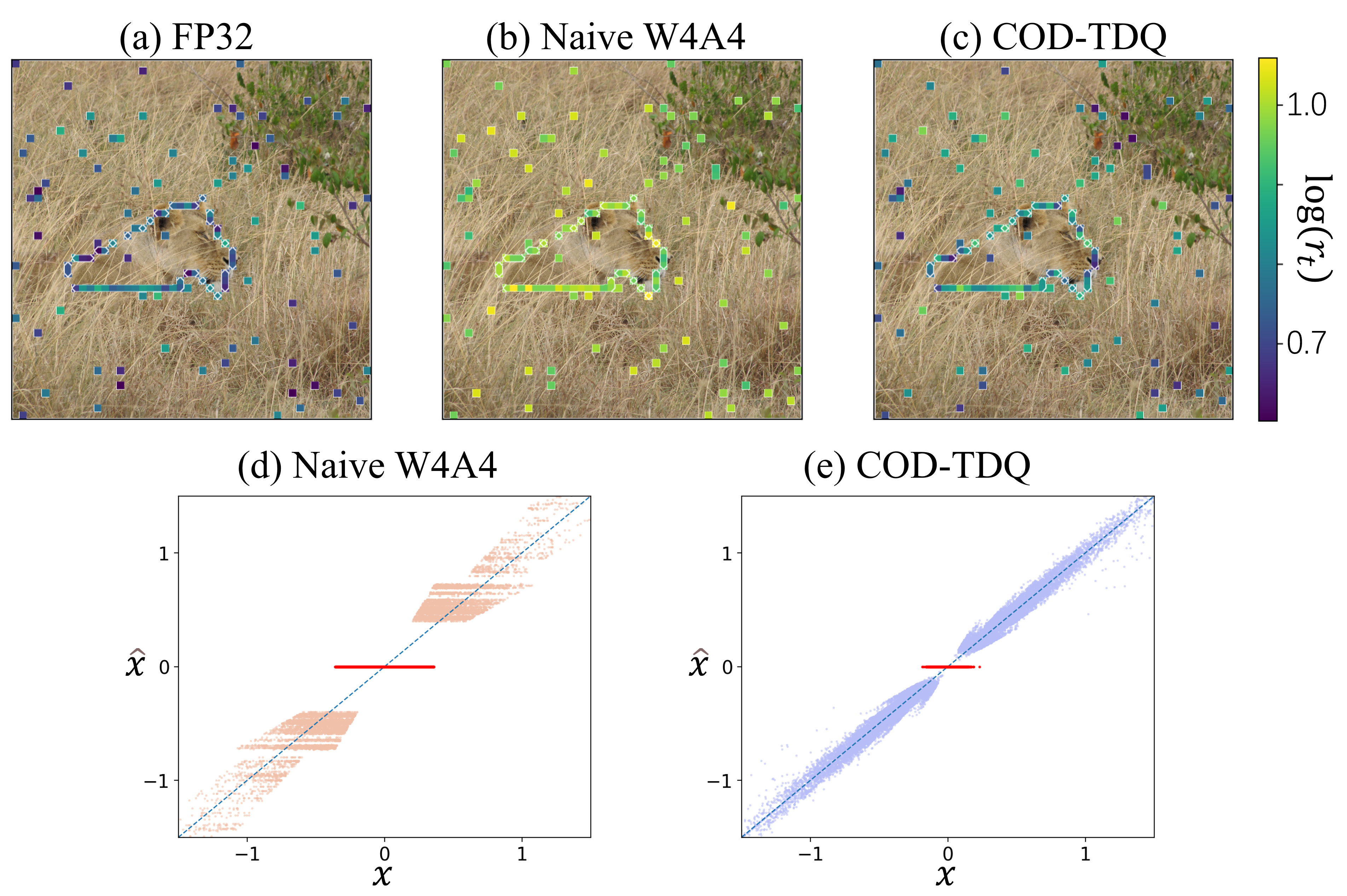}
\caption{\textbf{Reduces cross-token scale interference.} (a--c) Token-wise range disparity under FP32, naive W4A4, and DSTG: token-group scaling mitigates background-dominated range inflation. (d--e) Boundary-region activation magnitudes before/after quantization: naive W4A4 collapses many small responses to zero, while COD-TDQ preserves them, reducing the zeroed-activation fraction from 41.6\% to 14.2\%.}
\label{fig:DSTG_vis}
\end{figure*}

\subsection{Range Domination}
\label{sec:range_domination}
COD is intentionally low-contrast, so useful evidence often takes the form of small-amplitude but spatially structured responses.
At the same time, the token population is dominated by diverse backgrounds.
This imbalance creates strong token-wise activation heterogeneity.
A tensor-wise quantizer couples all tokens through a single clip radius $c$.
A small number of heavy-tailed background spikes can therefore dominate the range selection, increase $c$, and enlarge the step size $\Delta$ in~\cref{eq:cg_Delta_eta_oneline}.
Cross-token heterogeneity is summarized by the \textbf{range disparity}
\begin{equation}
\mathcal{D}(X)=\frac{\max_{c} |x_{c}|}{\operatorname{median}_{t}\ \operatorname{median}_{c}\ |x_{c}|},
\label{eq:range_disparity}
\end{equation}
which becomes large when a small fraction of tokens carries extreme magnitudes.
The shared range is set by outliers rather than by the majority of tokens, so most tokens are quantized with an overly coarse resolution. W4A4 amplifies the token-wise range disparity $\mathcal{D}(X)$ due to background spikes (\cref{fig:DSTG_vis}), whereas token-group scaling (\cref{sec:DSTG}) markedly suppresses the cross-token range inflation.

\begin{figure*}[!t]
\centering
\includegraphics[width=\textwidth]{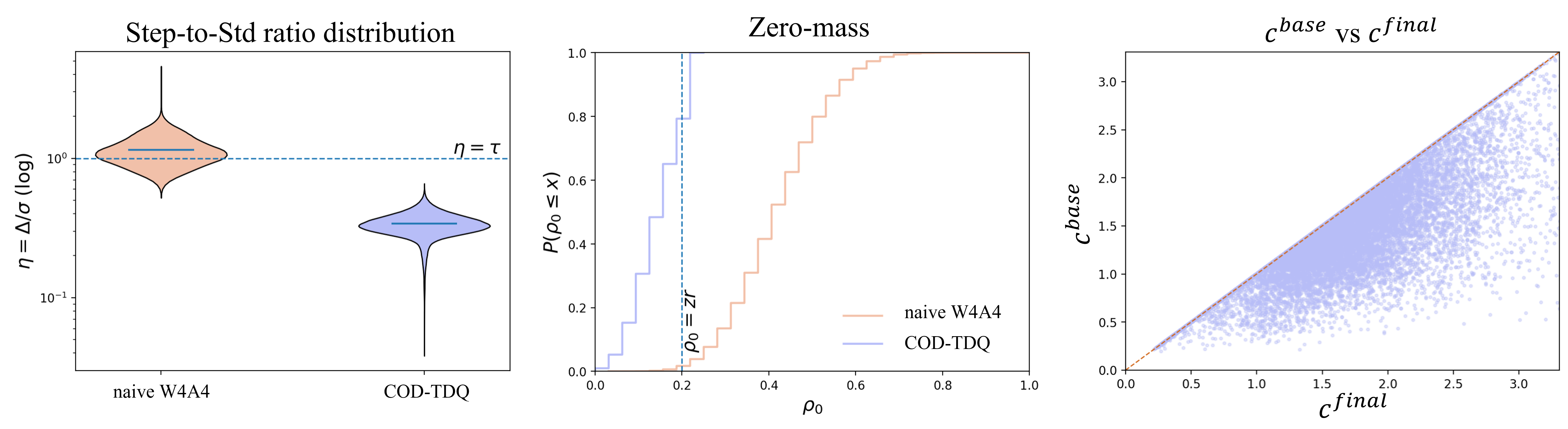}
\caption{\textbf{DCRP prevents zero-bin mass collapse.} DCRP projects each token-group clip radius to satisfy a step-to-dispersion bound and a zero-bin mass bound. The fraction of non-boundary token-groups exceeding the step-to-std threshold drops from 72.60\% (pre-projection) to 0.00\% after C1, and the fraction with pre-projection \(\rho_0>\mathrm{zr}\) drops from 98.36\% (naive W4A4) to 20.87\% under COD-TDQ statistics.}
\label{fig:dcrp_vis}
\end{figure*}

\subsection{Zero-bin Mass Collapse}
\label{sec:zero_mass_collapse}
Coarse step size becomes catastrophic in COD by increasing the zero-bin mass on boundary-heavy activations.
Range domination becomes destructive in COD because the task depends on weak boundary cues.
Once the global step size becomes coarse, many of these cues fall into the zero bin and disappear.

Under rounding-to-nearest, $\hat{x}=0$ holds when $|x|\le \Delta/2$.
The zero-bin mass in~\cref{eq:rho0_def} therefore increases monotonically with $\Delta$ for any fixed activation distribution.
For boundary-heavy activations, the increase is often steep because boundary responses cluster near zero.
After zeroization, subsequent attention mixing and linear projections cannot recreate the missing signed evidence, since the input to those operations is exactly zero.
This creates a token-local bottleneck that manifests as a global mask failure. Naive W4A4 places many token-groups in an unstable regime with overly large step-to-dispersion ratio $\eta$ and excessive zero-bin mass $\rho_0$, motivating explicit control of both diagnostics (\cref{fig:dcrp_vis}).

\para{Evidence from Mechanism Diagnostics.}
\label{sec:exp_motivation}
\label{sec:failure_evidence}
The diagnosis is linked to measurable forward-pass signals and to the representative numbers reported in~\cref{fig:failure_mechanism}.
The coupled range-domination and zero-bin mass loop is measurable during forward passes. \cref{fig:failure_mechanism} reports representative diagnostics on CFRN evaluated on NC4K.
Naive W4A4 increases the failure signals together with the accuracy collapse.
The bit-width diagnostic in~\cref{tab:motivation} supports this view.

Naive W8A8 and W4A8 remain close to FP32, while naive W4A4 collapses sharply.
Specifically, \cref{tab:motivation} shows that $S_\alpha$ drops from $0.888$ (FP32) to $0.443$ (naive W4A4), while the global clip factor increases from $c_g=2.794$ to $5.121$, the step size inflates from $\Delta=0.404$ to $3.651$, and the zero-bin mass rises from $\rho_0=0.141$ to $0.418$.
The same figure shows that COD-TDQ brings the diagnostics back toward the FP32 regime, with $S_\alpha=0.837$, $c_g=3.114$, $\Delta=0.565$, and $\rho_0=0.191$ under W4A4, as shown in~\cref{fig:failure_mechanism}.

These measurements isolate two requirements for reliable W4A4 COD.
First, range selection must be token-local to prevent background tokens from dominating the dynamic range.
Second, the quantizer must explicitly control zeroization for boundary-heavy activations.
~\cref{sec:method} operationalizes these requirements with token-group ranges (DSTG) and dual-constraint range projection (DCRP).
Detailed testable predictions and measurement protocols are provided in Supp. Sec.~S2.5.
The dominant failure factor is therefore 4-bit activations rather than 4-bit weights.

Static Symmetric Weight Quantization.
Weights are quantized once with standard symmetric uniform quantization and remain fixed during inference.
Activation robustness is dominated by the activation-side design in~\cref{sec:DSTG}.
Let $W\in\mathbb{R}^{O\times I}$ denote a weight matrix with output dimension $O$ and input dimension $I$.
Let $s\in\mathbb{R}^{O}$ denote per-output-channel scales with $s_o>0$.
We use broadcasted division: $(W/s)_{o,i}=W_{o,i}/s_o$.
Weight quantization is
\begin{equation}
Q_w(W)=\operatorname{clip}\big(\lfloor W/s \rceil,\ q_{\min}^w,q_{\max}^w\big),\qquad \hat{W}=s\odot Q_w(W),
\label{eq:weight_quant}
\end{equation}
where $\odot$ denotes element-wise multiplication with broadcasting along $I$.
Packing and storage are implementation details and are summarized in Supp. Sec.~S1.2.

\section{Method}
\label{sec:method}
This section specifies post-training simulated quantization under W4A4 and details COD-TDQ, a token-local activation quantization framework for Transformer-based camouflaged object detection. The design follows the fragility diagnosis in~\cref{sec:cod_failure} by suppressing cross-token range domination and by explicitly controlling the 4-bit step size and the induced zero-bin mass.

\begin{figure}[t]
\centering
\includegraphics[width=0.95\linewidth]{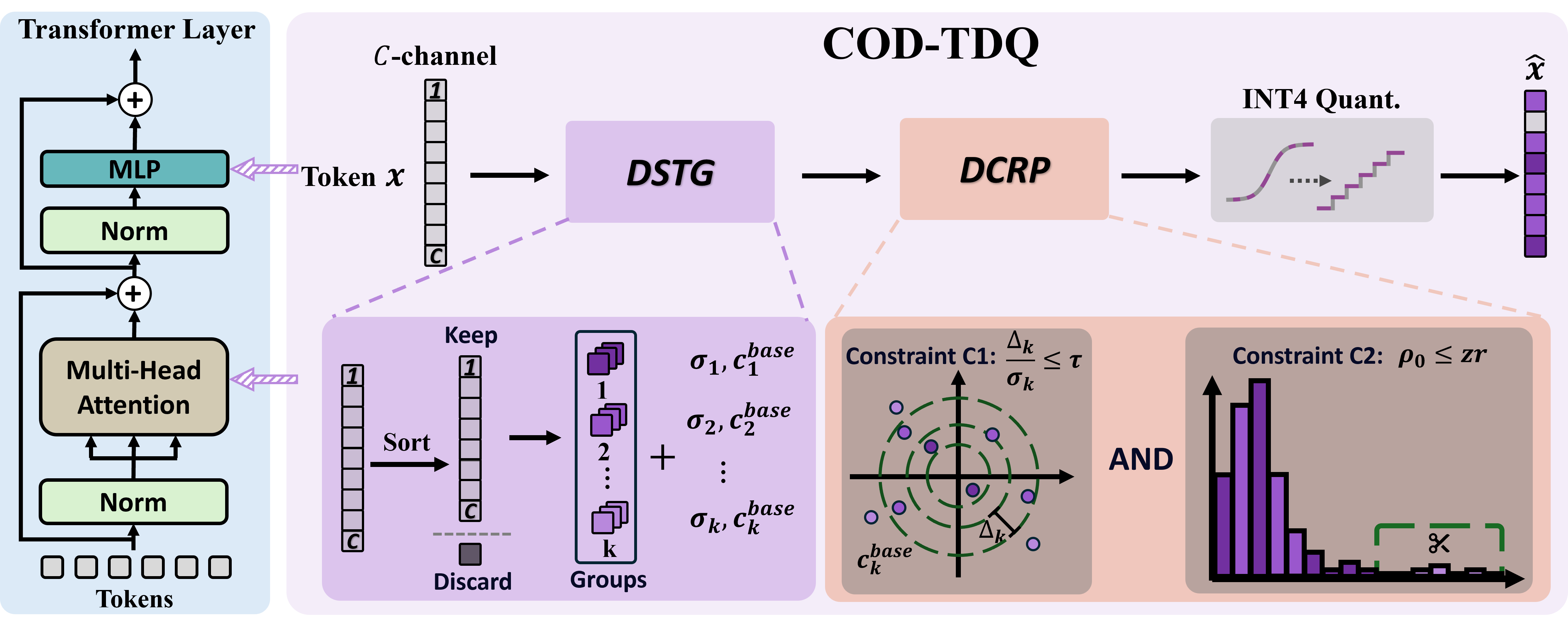}
\caption{\textbf{COD-TDQ (DSTG and DCRP) pipeline overview}. 
}
\label{fig:framework}
\end{figure}
\subsection{Overview of COD-TDQ}
\label{sec:overview}
Guided by the fragility diagnosis in~\cref{sec:cod_failure}, COD-TDQ treats COD quantization stability as a token-local range selection problem, aiming to suppress cross-token \textbf{range domination} and to explicitly control the 4-bit step size and the induced \textbf{zero-bin mass}.
To this end, COD-TDQ introduces two coupled modules---\textbf{DSTG} and \textbf{DCRP}---that operate at the token-group granularity and are detailed in~\cref{sec:DSTG} and~\cref{sec:dcrp}, (\cref{fig:framework}). Forward pseudocode is given in the supplements. \cref{fig:DSTG_vis}(d--e) shows the practical payoff of this token-local design: compared to naive W4A4, COD-TDQ preserves many weak boundary activations that would otherwise be rounded into the zero bin.

Symmetric uniform quantization. The operator $\operatorname{clip}(\cdot)$ clips element-wise to a closed interval.
The notation $\lfloor\cdot\rceil$ denotes rounding to the nearest integer. Given a clip radius $c>0$, the step size is $\Delta=c/q_{\max}$.
A scalar $x$ is clipped to $\tilde{x}=\operatorname{clip}(x,-c,c)$, quantized as $q=\operatorname{clip}(\lfloor \tilde{x}/\Delta\rceil,q_{\min},q_{\max})$, and dequantized as $\hat{x}=\Delta q$.
A small constant $\varepsilon>0$ avoids degenerate steps in implementation.
COD-TDQ targets the dominant projection operators, namely \texttt{Linear} layers in attention and MLP blocks and \texttt{Conv} layers when present.
Full operator coverage and implementation details are listed in Supp. Sec.~S1.1.

COD-TDQ instantiates token-local range selection with two coupled modules.
DSTG localizes scaling to token groups to remove cross-token range domination.
DCRP then projects each group range so that discretization and zeroization remain bounded under 4-bit activations.
\textbf{DSTG} (\textbf{D}irect-\textbf{S}um \textbf{T}oken-\textbf{G}roup) partitions each token vector into fixed-size channel groups and assigns each group its own activation range.
\textbf{DCRP} (\textbf{D}ual-\textbf{C}onstraint \textbf{R}ange \textbf{P}rojection) adjusts each group range with two constraints that control the step-to-dispersion ratio and the zero-bin mass.
We next detail DSTG in~\cref{sec:DSTG}, and then introduce DCRP in~\cref{sec:dcrp} to complete the token-group W4A4 quantizer.

\subsection{Direct-Sum Token-Group}
\label{sec:DSTG}
Token-wise activation heterogeneity in COD lets heavy-tailed background tokens dominate this shared range, inflating the step size and increasing the zero-bin mass on boundary-heavy activations, as discussed in \cref{sec:cod_failure}.
This cross-token coupling motivates a token-local range assignment that decouples background-driven outliers from the majority of tokens.
In response, DSTG assigns a dedicated activation range to each token-group, which prevents background tokens from dictating a shared clipping range.
DSTG decomposes each token into groups and performs uniform quantization.

Direct-Sum Token-Group Decomposition. For a token vector $x \in\mathbb{R}^{C}$, channels are grouped into blocks of size $g$.
If $C$ is not divisible by $g$, the vector is zero-padded on the channel dimension to $C_{\mathrm{pad}}=g\lceil C/g\rceil$.
\begin{equation}
x=\bigoplus_{k=1}^{K} x_{k},\qquad x_{k}\in\mathbb{R}^{g},\qquad K=\frac{C_{\mathrm{pad}}}{g},
\label{eq:direct_sum}
\end{equation}
\begin{equation}
c^{\text{base}}_{k}=
\begin{cases}
\|x_{k}\|_{\infty}, & \text{without percentile clipping},\\
Q_{p}\!\big(|x_{k}|\big), & \text{with percentile }p\in(0,1],
\end{cases}
\label{eq:DSTG_clip}
\end{equation}
The padded token is decomposed as~\cref{eq:direct_sum}. Where $\oplus$ denotes concatenation along the channel dimension.
Padding is removed after dequantization.
A base clip radius $c^{\text{base}}_{k}>0$ is estimated from the magnitudes of the group vector $x_{k}$ as~\cref{eq:DSTG_clip}. Where $\|\cdot\|_{\infty}$ is the max-absolute norm and $Q_{p}$ is the empirical $p$-quantile over the $g$ elements of $|x_{k}|$.
Token-group scaling plays an important role. Let $c^{\text{global}}=\max_{k} c^{\text{base}}_{k}$ denote a per-tensor radius used by a shared-range quantizer.
Let $\sigma_{k}=\operatorname{Std}(x_{k})+\varepsilon$ denote the within-group standard deviation, computed over the $g$ elements.
The corresponding step-to-dispersion ratio under a shared range satisfies
\begin{equation}
\frac{c^{\text{global}}}{q_{\max}\sigma_{k}}
\gg
\frac{c^{\text{base}}_{k}}{q_{\max}\sigma_{k}}
\label{eq:interference}
\end{equation}
whenever $c^{\text{global}}$ is dominated by outliers from other tokens or groups.
Such cross-token interference is frequent in COD and motivates token-group ranges.

Uniform signed quantization per token-group.
Given the final clip radius \(c_k\) after DCRP (\cref{sec:dcrp}), DSTG applies signed uniform quantization within the group:
\begin{align}
\Delta_{k} &= \max\!\left(\frac{c_{k}}{q_{\max}},\ \varepsilon\right), & 
\tilde{x}_{k} &= \operatorname{clip}(x_{k},-c_{k},c_{k}), \label{eq:clipx}\\
q_{k} &= \operatorname{clip}\Big(\big\lfloor \tilde{x}_{k}/\Delta_{k}\big\rceil,\ q_{\min},q_{\max}\Big)\in\mathbb{Z}^{g}, & 
\hat{x}_{k} &= \Delta_{k}\, q_{k}. \label{eq:dequant}
\end{align}
The quantized token is reconstructed as \(\hat{x}_t=\bigoplus_k \hat{x}_k\), and the padded dimensions are removed to recover \(C\) channels.

\subsection{Dual-Constraint Range Projection}
\label{sec:dcrp}
DSTG removes cross-token coupling, but a single token-group is still heavy-tailed and inflates its own range.
DCRP therefore projects each group radius to satisfy two stability constraints that directly control discretization and zeroization.

\noindent\textbf{Constraint C1: step-to-dispersion bound.}
The first constraint upper-bounds the step-to-dispersion ratio by a user-specified $\tau>0$ ($\sigma_{k}=\operatorname{Std}(x_{k})+\varepsilon$).
\begin{equation}
\eta_{k}\triangleq \frac{\Delta_{k}}{\sigma_{k}} \le \tau
\quad \Longleftrightarrow \quad
c_{k} \le c^{(\tau)}_{k}\triangleq q_{\max}\tau\sigma_{k}.
\label{eq:tau_constraint}
\end{equation}

\noindent\textbf{Constraint C2: zero-bin mass bound.}
Under rounding-to-nearest, an element quantizes to zero if $|x|\le \Delta/2$.
The empirical zero-bin mass is
\begin{equation}
\rho_{0,k}\triangleq \frac{1}{g}\sum_{i=1}^{g}\mathbf{1}\!\left(|x_{k,i}|\le \frac{\Delta_{k}}{2}\right),\quad
c_{k} \le c^{(\mathrm{zr})}_{k}\triangleq 2q_{\max}\,Q_{\mathrm{zr}}\!\big(|x_{k}|\big).
\label{eq:zr_constraint}
\end{equation}

A target bound $\mathrm{zr}\in(0,1)$ limits zeroization by enforcing $\rho_{0,k}\le \mathrm{zr}$.
Let $Q_{\mathrm{zr}}(|x_{k}|)$ denote the empirical $\mathrm{zr}$-quantile of the $g$ magnitudes in the group. 
The condition $\rho_{0,k}\le \mathrm{zr}$ is satisfied when $\Delta_{k}/2 \le Q_{\mathrm{zr}}(|x_{k}|)$, which yields the bound on $c_k$ shown above.
The feasible interval for the clip radius and the corresponding projection are
\begin{equation}
\mathcal{C}_{k}=\Big(0,\ \min\{c^{(\tau)}_{k},\ c^{(\mathrm{zr})}_{k}\}\Big],\qquad
c_{k}=\Pi_{\mathcal{C}_{k}}\!\left(c^{\text{base}}_{k}\right)
=\min\!\left(c^{\text{base}}_{k},\ c^{(\tau)}_{k},\ c^{(\mathrm{zr})}_{k}\right).
\label{eq:dcrp}
\end{equation}
We then set \(c_k \leftarrow \max(c_k,\varepsilon)\).
By construction, the projected radius satisfies
\begin{equation}
\eta_{k}\le \tau,\qquad \rho_{0,k}\le \mathrm{zr}
\quad \text{(up to empirical-quantile discretization)}.
\label{eq:diagnostics_as_constraints}
\end{equation}
As illustrated in~\cref{fig:dcrp_vis}, the projection in~\cref{eq:dcrp} sharply reduces the fraction of token-groups violating the C1/C2 bounds on $(\eta_k,\rho_{0,k})$, preventing zero-bin mass collapse in practice.
The clipping and rounding trade-off, together with rounding-noise bounds implied by Constraint C1, are provided in Supp. Sec.~S2.3.

Putting the two modules together.
DSTG removes cross-token range domination by assigning token-group ranges (\cref{eq:interference}).
DCRP prevents each group from drifting into an unstable W4A4 regime where the step is too coarse or where zeroization is excessive (\cref{eq:diagnostics_as_constraints}).
This combination directly targets the failure loop analyzed in~\cref{sec:cod_failure} and is validated by diagnostics in~\cref{sec:exp_qual}.

\section{Experiments}
\label{sec:experiments}

\begin{table*}[t]
\centering
\caption{\textbf{W4A4 post-training quantization results on CFRN.}}
\label{tab:cfrn_w4a4}
\scriptsize
\setlength{\tabcolsep}{2.8pt}
\renewcommand{\arraystretch}{1.12}
\resizebox{\linewidth}{!}{
\begin{tabular}{l | ccccc | ccccc | ccccc | ccccc}
\toprule
\multirow{2}{*}{Method} 
& \multicolumn{5}{c|}{\textbf{CAMO}} 
& \multicolumn{5}{c|}{\textbf{CHAMELEON}}
& \multicolumn{5}{c|}{\textbf{COD10K}}
& \multicolumn{5}{c}{\textbf{NC4K}} \\
\cmidrule(lr){2-6}
\cmidrule(lr){7-11}
\cmidrule(lr){12-16}
\cmidrule(lr){17-21}
& $S_\alpha$ & $F^{\omega}_{\beta}$ & $E_m$ & $F^m_\beta$ & MAE
& $S_\alpha$ & $F^{\omega}_{\beta}$ & $E_m$ & $F^m_\beta$ & MAE
& $S_\alpha$ & $F^{\omega}_{\beta}$ & $E_m$ & $F^m_\beta$ & MAE
& $S_\alpha$ & $F^{\omega}_{\beta}$ & $E_m$ & $F^m_\beta$ & MAE \\
\midrule

\multicolumn{21}{c}{\textit{Full Precision \& Naive Quantization}} \\
\midrule

FP32 
& .876 & .844 & .934 & .877 & .042
& .912 & .875 & .961 & .898 & .019
& .870 & .795 & .939 & .835 & .022
& .888 & .850 & .941 & .880 & .030 \\

Naive W8A8
& .875 & .843 & .933 & .876 & .042
& .912 & .875 & .961 & .897 & .019
& .869 & .795 & .938 & .834 & .022
& .887 & .850 & .940 & .879 & .030 \\

Naive W4A8
& .864 & .826 & .923 & .861 & .047
& .906 & .863 & .959 & .886 & .021
& .858 & .776 & .931 & .818 & .024
& .882 & .841 & .936 & .872 & .032 \\

Naive W4A4
& .418 & .061 & .314 & .332 & .182
& .447 & .083 & .351 & .316 & .141
& .471 & .071 & .376 & .242 & .093
& .443 & .089 & .344 & .348 & .151 \\

\midrule
\multicolumn{21}{c}{\textit{Existing PTQ Methods}} \\
\midrule

NoisyQuant
& .408 & .112 & .498 & .331 & .190
& .443 & .120 & .574 & .318 & .150
& .474 & .101 & .380 & .248 & .095
& .450 & .173 & .324 & .341 & .124 \\

PTQ4ViT
& .410 & .061 & .352 & .318 & .181
& .440 & .080 & .380 & .321 & .140
& .479 & .070 & .393 & .252 & .091
& .456 & .081 & .340 & .355 & .153 \\

ORQ-ViT
& .415 & .069 & .303 & .321 & .181
& .448 & .083 & .356 & .316 & .141
& .489 & .086 & .364 & .240 & .092
& .440 & .081 & .332 & .344 & .151 \\

post-GELU
& .403 & .093 & .469 & .319 & .185
& .449 & .084 & .466 & .318 & .147
& .480 & .073 & .544 & .259 & .084
& .450 & .094 & .502 & .352 & .156 \\

SmoothQuant
& .397 & .118 & .526 & .321 & .140
& .445 & .104 & .558 & .328 & .140
& .490 & .073 & .565 & .249 & .091
& .452 & .108 & .543 & .342 & .157 \\

PQ-SAM
& .464 & .228 & .511 & .398 & .121
& .465 & .190 & .499 & .317 & .150
& .483 & .131 & .462 & .243 & .089
& .474 & .211 & .508 & .351 & .154 \\

QuaRTZ
& .416 & .054 & .309 & .324 & .182
& .447 & .083 & .353 & .319 & .140
& .471 & .070 & .371 & .243 & .093
& .441 & .084 & .337 & .350 & .151 \\

AHCPTQ
& .417 & .060 & .313 & .329 & .182
& .449 & .086 & .364 & .320 & .141
& .471 & .070 & .374 & .243 & .093
& .442 & .086 & .340 & .345 & .151 \\

ARCQuant
& .534 & .510 & .657 & .629 & .117
& .593 & .464 & .619 & .572 & .096
& .640 & .455 & .676 & .570 & .079
& .683 & .569 & .711 & .674 & .097 \\

FIMA-Q
& .595 & .549 & .659 & .553 & .130
& .603 & .476 & .682 & .551 & .096
& .611 & .452 & .698 & .518 & .076
& .663 & .572 & .733 & .629 & .089 \\

PTQ4SAM
& .671 & .591 & .740 & .653 & .106
& .605 & .461 & .645 & .518 & .095
& .665 & .540 & .760 & .584 & .072
& .702 & .632 & .761 & .660 & .082 \\

IGQ-ViT
& .674 & .601 & .748 & \underline{.658} & .113
& .692 & .607 & .763 & .659 & .069
& .670 & .538 & .778 & .592 & .069
& .710 & \underline{.651} & .776 & .688 & .080 \\

RepQ-ViT
& \underline{.676} & \underline{.608} & \underline{.750} & .656 & \underline{.099}
& \underline{.701} & \underline{.622} & \underline{.774} & \underline{.672} & \underline{.067}
& \underline{.676} & \underline{.549} & \underline{.779} & \underline{.599} & \underline{.062}
& \underline{.718} & \underline{.651} & \underline{.783} & \underline{.691} & \underline{.072} \\

\midrule
\textbf{Ours}
& \textbf{.813} & \textbf{.724} & \textbf{.864} & \textbf{.795} & \textbf{.070}
& \textbf{.862} & \textbf{.758} & \textbf{.904} & \textbf{.823} & \textbf{.040}
& \textbf{.802} & \textbf{.649} & \textbf{.864} & \textbf{.736} & \textbf{.038}
& \textbf{.837} & \textbf{.747} & \textbf{.884} & \textbf{.817} & \textbf{.052} \\

\bottomrule
\end{tabular}
}
\end{table*}

\subsection{Experimental setup}
\label{sec:exp_setup}

\para{Datasets.}
We evaluate on four standard COD benchmarks: {CAMO}~\cite{le2019anabranch} ($1000$ train / $250$ test), {CHAMELEON}~\cite{chameleon} ($76$ images), {COD10K}~\cite{fan2020cod} (test split with $2026$ images), and {NC4K}~\cite{lv2021rank} ($4121$ images).
Unless stated otherwise, we report results on the official test splits using the original image resolutions and the evaluation protocols released by prior COD work.
Following COD conventions \cite{fan2020cod}, we report five metrics: $S_\alpha$ (structure measure), $F^{\omega}_{\beta}$ (weighted F-measure), $E_m$ (mean E-measure), $F^m_\beta$ (max F-measure), and MAE.
For $S_\alpha,F^{\omega}_{\beta},E_m,F^m_\beta$, higher is better, while lower MAE indicates better mask quality.

\para{Models.}
We primarily study {CFRN}~\cite{song2025cfrn}, a strong Swin-based COD model with Transformer encoder blocks and a COD-specific decoder.
We also evaluate on {ESCNet}~\cite{ye2025escnet}, a Pyramid Vision Transformer (PVT)-style Transformer COD model with edge/texture collaboration modules.

\para{Quantization protocol.}
We focus on \textbf{W4A4}, and use W8A8/W4A8 only as diagnostic references to localize the failure source.
Weights are quantized once with symmetric uniform quantization (\cref{sec:failure_prelim}).
All comparisons are accuracy-centric under the same evaluation pipeline.
In all experiments, we apply COD-TDQ to \texttt{Linear}/\texttt{Conv} operators, while quantizing \texttt{LayerNorm}/\texttt{Softmax} with a conventional FP16 routine.
We use a single hyperparameter setting shared across all datasets and both CFRN/ESCNet baselines: $g=32$, $\tau=1.0$, and $\mathrm{zr}=0.2$.

\para{Baselines and reproduction.}
We compare COD-TDQ against representative Transformer PTQ methods and cross-domain PTQ transfers listed in~\cref{tab:cfrn_w4a4} and \cref{tab:escnet_nc4k}.
For baselines that require offline calibration, we use a 128-image calibration set.
All results are obtained under the same evaluation codebase with identical preprocessing and post-processing as the original FP32 models.
All quantization methods operate on the same set of quantized layers for fairness.

\subsection{Main results on CFRN}
\label{sec:exp_cfrn}

\begin{table*}[t]
\centering
\caption{\textbf{W4A4 post-training quantization on ESCNet.} We bold the best results and underline the second-best results, including ties, among all PTQ methods.}
\label{tab:escnet_nc4k}
\scriptsize

\setlength{\tabcolsep}{4.5pt} 
\renewcommand{\arraystretch}{1.2}  

\resizebox{\linewidth}{!}{
\begin{tabular}{l | ccc | ccccccccc | c}
\toprule
\multirow{2}{*}{\textbf{Method}} & \multicolumn{3}{c|}{\textbf{Baselines}} & \multicolumn{9}{c|}{\textbf{Existing PTQ Methods}} & \textbf{Ours} \\
\cmidrule(lr){2-4} \cmidrule(lr){5-13} \cmidrule(lr){14-14}
 & FP32 & W8A8 & W4A8 & ORQ & GELU & PQ & Noisy & PTQ4V & FIMA & PTQ4S & IGQ & RepQ & \textbf{COD-TDQ} \\
\midrule
$S_\alpha\uparrow$    & .893 & .893 & .888 & .576 & .581 & .609 & .714 & .723 & .748 & .768 & \underline{.818} & \underline{.818} & \textbf{.881} \\
$F^{\omega}_{\beta}\uparrow$   & .864 & .864 & .858 & .368 & .375 & .429 & .651 & .658 & .662 & .701 & .727 & \underline{.751} & \textbf{.849} \\
$E_m\uparrow$        & .945 & .945 & .941 & .562 & .566 & .607 & .803 & .818 & .821 & .843 & .878 & \underline{.883} & \textbf{.936} \\
$F^m_\beta\uparrow$  & .887 & .887 & .883 & .459 & .467 & .526 & .684 & .700 & .710 & .741 & .790 & \underline{.791} & \textbf{.876} \\
MAE$\downarrow$      & .028 & .028 & .029 & .120 & .119 & .111 & .093 & .080 & .078 & .062 & .055 & \underline{.052} & \textbf{.031} \\
\bottomrule
\end{tabular}
}
\end{table*}

\begin{table}[t]
\centering
\caption{\textbf{Ablation studies on NC4K dataset.} Comparison of components on CFRN (left) and ESCNet (right) baselines. Best results are in \textbf{bold}.}
\label{tab:ablation_aligned}
\tiny

\setlength{\tabcolsep}{3pt}  
\renewcommand{\arraystretch}{1.3}

\begin{minipage}[t]{0.48\linewidth}
\centering
\begin{tabular}{l | ccccc}
\toprule
\multirow{2.5}{*}{\textbf{Method}} & \multicolumn{5}{c}{\textbf{CFRN Backbone}} \\
\cmidrule(lr){2-6}
 & $S_\alpha\uparrow$ & $F^{\omega}_{\beta}\uparrow$ & $E_m\uparrow$ & $F^m_\beta\uparrow$ & MAE$\downarrow$ \\
\midrule
Naive W4A4 & .443 & .089 & .344 & .348 & .151 \\
Per-tensor & .407 & .093 & .513 & .185 & .209 \\
DSTG only  & .451 & .180 & .532 & .241 & .270 \\
DCRP only  & .435 & .174 & .504 & .228 & .307 \\
\midrule
\textbf{Ours} & \textbf{.837} & \textbf{.747} & \textbf{.884} & \textbf{.817} & \textbf{.052} \\
\bottomrule
\end{tabular}
\end{minipage}
\hfill
\begin{minipage}[t]{0.48\linewidth}
\centering
\begin{tabular}{l | ccccc}
\toprule
\multirow{2.5}{*}{\textbf{Method}} & \multicolumn{5}{c}{\textbf{ESCNet Backbone}} \\
\cmidrule(lr){2-6}
 & $S_\alpha\uparrow$ & $F^{\omega}_{\beta}\uparrow$ & $E_m\uparrow$ & $F^m_\beta\uparrow$ & MAE$\downarrow$ \\
\midrule
Naive W4A4 & .576 & .368 & .562 & .459 & .120 \\
Per-tensor & .597 & .378 & .548 & .448 & .111 \\
DSTG only  & .617 & .416 & .579 & .488 & .106 \\
DCRP only  & .771 & .684 & .805 & .744 & .062 \\
\midrule
\textbf{Ours} & \textbf{.881} & \textbf{.849} & \textbf{.936} & \textbf{.876} & \textbf{.031} \\
\bottomrule
\end{tabular}
\end{minipage}
\end{table}

Bit-width sensitivity supports the COD-specific failure diagnosis.
Across all four datasets, \textit{naive W4A8} stays close to FP32 (average $S_\alpha$: $0.8776$ vs.\ $0.8864$ for FP32), while \textit{naive W4A4} collapses severely (average $S_\alpha$: $0.4448$, average MAE: $0.1417$).
This pattern localizes the dominant failure factor to {4-bit activations}, consistent with~\cref{sec:cod_failure}.
COD-TDQ achieves the best W4A4 accuracy on CFRN across all datasets (\cref{tab:cfrn_w4a4}).
Compared to the strongest W4A4 baselines, COD-TDQ improves $S_\alpha$ by \textbf{+11.8 to +16.1} points and reduces MAE by \textbf{0.008 to 0.020} absolute.
On {NC4K}, COD-TDQ reaches $S_\alpha=0.8365$ with MAE $0.0520$, while the best baseline (RepQ-ViT~\cite{li2023repqvit}) attains $S_\alpha=0.7182$ with MAE $0.0715$.

{PTQ4ViT} \cite{yuan2022ptq4vit} targets ViT quantization with twin uniform quantization and Hessian-guided scale selection, but its shared scale assumption is brittle under COD token heterogeneity.
{FIMA-Q} \cite{wu2025fimaq} and {RepQ-ViT}~\cite{li2023repqvit} improve reconstruction fidelity, yet remain layer/block-centric and do not explicitly bound token-local $(\eta,\rho_0)$, leaving a consistent gap to COD-TDQ.
Channel grouping or bit-width allocation.
{IGQ-ViT} \cite{moon2024igqvit} alleviates channel-level outliers but does not remove cross-token interference, hence remains limited by token-local zeroization.
{post-GELU} \cite{kim2026tokenbitwidth} assigns dynamic bit-widths, but without correcting token-wise scale mismatch or bounding $\rho_0$, it still fails under fixed W4A4.
Outlier-centric methods.
Outlier suppression or noise injection does not directly prevent boundary cues from collapsing into the zero bin when $\Delta$ is coarse (\cref{tab:ablation_aligned}).
Accordingly, ORQ-ViT~\cite{ning2025orqvit} and NoisyQuant~\cite{yang2023noisyquant} remain far from FP32 on CFRN, and SmoothQuant~\cite{xiao2023smoothquant} is mainly effective in higher-activation-bit regimes.
Transferred PTQ methods target different activation pathologies and do not explicitly intervene on COD token-wise heterogeneity and boundary-sensitive zeroization.
COD-TDQ differs by enforcing token-group constraints on~\cref{eq:diagnostics_as_constraints}.

\subsection{Transferability to ESCNet}
\label{sec:exp_escnet}
COD-TDQ generalizes across Transformer backbones.
On ESCNet, naive W4A4 also degrades sharply.
COD-TDQ restores ESCNet to near-lossless W4A4 performance.
Full W4A4 tables on all datasets are in Supp. Sec.~S3.3.
RepQ-ViT and IGQ-ViT remain strong, but their channel-wise mechanisms do not prevent token-local zeroization.
In contrast, COD-TDQ applies token-group constraints.

On CFRN, removing token-local scaling (Per-tensor) exposes severe cross-token range domination and does not recover W4A4 performance.
DSTG only also fails when local clip radii are still inflated by tails, consistent with the zero-bin mass collapse diagnosis (\cref{sec:cod_failure}).
DCRP only in attention yields limited gains because unstable ranges in MLP projections can still erase boundary evidence.
Only DSTG + DCRP consistently restores COD masks.
On ESCNet, DCRP provides strong stabilization, and DSTG closes the remaining gap.
On ESCNet, DCRP only already recovers a large portion of W4A4 accuracy, confirming that bounding $\Delta/\sigma$ and $\rho_0$ targets the dominant error mode.
Adding DSTG further improves accuracy and reduces MAE, showing the complementarity of local scale alignment and constraint-based projection.

\subsection{Qualitative Analysis}
\label{sec:exp_qual}

We visualize COD-TDQ's two components with mechanism-level diagnostics: DSTG reduces cross-token scale interference (\cref{fig:DSTG_vis}) and DCRP enforces token-group stability bounds on \(\eta=\Delta/\sigma\) and \(\rho_0\) (\cref{fig:dcrp_vis}).

\begin{figure*}[!t]
\centering
\includegraphics[width=\textwidth]{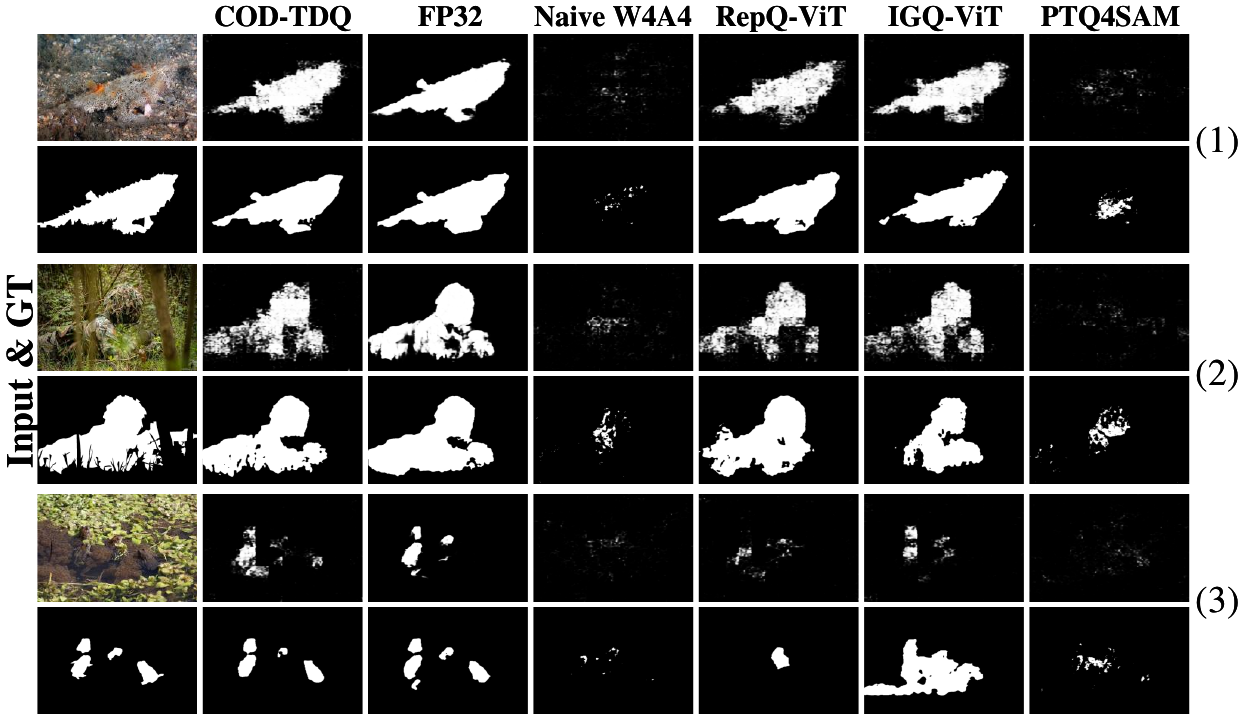}
\caption{\textbf{Qualitative comparison.}
The first column shows the input image and the GT mask. The remaining columns present the prediction masks produced by different quantization methods (RepQ-ViT, IGQ-ViT, PTQ4SAM). For each example, the two rows correspond to results obtained with the CFRN and ESCNet baselines, respectively.}
\label{fig:qualitative}
\end{figure*}

\cref{fig:qualitative} compares representative masks on challenging scenes (weak boundaries, textured backgrounds).
Naive W4A4 often degenerates into nearly uniform predictions or fragmented responses.
Strong ViT PTQ baselines (RepQ-ViT, IGQ-ViT) partially recover coarse structure but still miss fine contours.
COD-TDQ produces masks closest to FP32, particularly on thin boundaries and low-contrast foreground regions.

\subsection{Real INT4 Deployment}
\label{sec:real_deployment}

We further benchmark end-to-end CFRN inference using a Triton-based INT4 path on a single NVIDIA RTX~4090. We use $384\times384$ inputs and batch size 4. Latency is averaged per image, and memory denotes peak allocated GPU memory.

\begin{table}[t]
\centering
\caption{Runtime efficiency comparison on CFRN.}
\label{tab:cfrn-runtime-efficiency}
\setlength{\tabcolsep}{3.5pt}
\renewcommand{\arraystretch}{1.08}
\resizebox{0.95\textwidth}{!}{%
\begin{tabular}{lcccc}
\toprule
\textbf{Method}
& \textbf{Weight (MB)}$\downarrow$
& \textbf{FPS (img/s)}$\uparrow$
& \textbf{Latency (ms/img)}$\downarrow$
& \textbf{Peak Memory (MB)}$\downarrow$ \\
\midrule
FP32
& 775.40 & 29.40 & 34.02 & 1421.31 \\
Naive W4A4
& 108.30 & 43.95 & 22.75 & 843.30 \\
PTQ4SAM
& \textbf{108.13} & 43.83 & 22.82 & 848.80 \\
RepQ-ViT
& 108.15 & 44.18 & 22.64 & 844.67 \\
IGQ-ViT
& 108.14 & 44.19 & 22.63 & 852.23 \\
\midrule
\textbf{COD-TDQ}
& 108.19 & \textbf{44.21} & \textbf{22.61} & \textbf{834.92} \\
\bottomrule
\end{tabular}%
}
\end{table}

As shown in~\cref{tab:cfrn-runtime-efficiency}, COD-TDQ reaches 44.21 FPS and 22.61 ms/image, corresponding to a $1.50\times$ throughput gain over FP32, while reducing the deployment artifact from 775.40 to 108.19 MB and peak memory from 1421.31 to 834.92 MB. Its runtime is also comparable to the other W4A4 methods. Relative to the implementation without DSTG (44.48 FPS), DSTG incurs only a 0.27 FPS (0.6\%) overhead. The Triton path covers 69.73\% of quantized operators. The ideal packed INT4 weights occupy 97,065,266 bytes. Scale tensors and indexing metadata add 326,632 and 4,364 bytes, respectively, giving only 0.341\% combined storage overhead. DSTG uses contiguous token-group/channel-group scale indices and requires neither sorting nor an additional activation pass.

\section{Conclusion}
Camouflaged object detection (COD) attains high accuracy, yet Transformer-based COD models remain costly for mobile and edge deployment. Post-training quantization (PTQ) with 4-bit weights and activations (W4A4) is appealing. However, COD exhibits a pronounced, task-specific accuracy cliff. We trace this degradation mainly to 4-bit activations: token-wise heterogeneity and heavy-tailed background tokens dominate a shared clipping range, enlarge the step size, and suppress weak but structured boundary cues.
To counter this mechanism without retraining, we propose COD-TDQ, integrating Direct-Sum Token-Group scaling (DSTG) with Dual-Constraint Range Projection (DCRP). COD-TDQ  bounds both the discretization strength and zero-bin mass per token group. 
We anticipate this work will foster deployment-friendly COD quantization and inspire further study of tighter constraints, decoder diagnosis, cross-architecture generalization, and low-cost adaptation.

\section*{Acknowledgements}
This work is supported by the National Natural Science Foundation of China (No. 62576176). The computational resources are supported by the Supercomputing Center of Nankai University (NKSC).

%
%
\bibliographystyle{splncs04}
\bibliography{main}

\appendix

\title{When W4A4 Breaks Camouflaged Object Detection: Token-Group Dual-Constraint Activation Quantization
\texorpdfstring{\\}{ }Supplementary Material}

\author{Tianqi Li\inst{1}\orcidlink{0009-0007-5978-0389} \and
Wenyu Fang\inst{1}\orcidlink{0009-0005-8847-3017} \and
Xin He\inst{2}\orcidlink{0009-0004-2139-6590} \and
Xue Geng\inst{3}\orcidlink{0000-0002-2594-9648} \and
Xu Cheng\inst{2}\orcidlink{0000-0002-4724-5748} \and 
Yun Liu\inst{1,4,5}\thanks{Corresponding author: Yun Liu (liuyun@nankai.edu.cn)}\orcidlink{0000-0001-6143-0264}}

\authorrunning{Li et al.}


\institute{VCIP, College of Computer Science, Nankai University \and
School of Computer Science and Engineering, Tianjin University of Technology \and
Institute for Infocomm Research, A*STAR \and
Academy for Advanced Interdisciplinary Studies, Nankai University \and
Nankai International Advanced Research Institute, Shenzhen Futian}

\maketitle

\renewcommand{\thesection}{S\arabic{section}}
\renewcommand{\thesubsection}{S\arabic{section}.\arabic{subsection}}
\renewcommand{\thefigure}{S\arabic{figure}}
\renewcommand{\thetable}{S\arabic{table}}
\renewcommand{\thealgorithm}{S\arabic{algorithm}}
\renewcommand{\theequation}{S\arabic{equation}}
\setcounter{section}{0}
\setcounter{figure}{0}
\setcounter{table}{0}
\setcounter{algorithm}{0}
\setcounter{equation}{0}

\section{Implementation Details}
\label{sec:supp_impl}

\subsection{Quantized Operators}
\label{sec:supp_where}

COD-TDQ is implemented through wrapper replacement in the quantization builder.
When quantization is enabled, the builder replaces \texttt{nn.Linear}, \texttt{nn.Conv2d}, and \texttt{nn.Conv1d} with \texttt{QuantLinear}, \texttt{QuantConv2d}, and \texttt{QuantConv1d}, respectively.
The builder also supports optional module-pattern controls, including skip rules, mixed-precision exceptions, operator-specific group-size overrides, optional weight-clipping percentiles, and an optional per-layer JSON override stage applied after wrapper injection.
\texttt{LayerNorm} and \texttt{GroupNorm} remain on the floating-point path and can be explicitly wrapped to FP32 when \texttt{fp32\_ln} or \texttt{fp32\_gn} is enabled.

For \texttt{QuantLinear}, the activation path supports three modes in the codebase, namely \texttt{per\_tensor}, \texttt{per\_channel}, and \texttt{per\_token\_group}.
COD-TDQ uses the \texttt{per\_token\_group} path.
Given an input activation tensor $X\in\mathbb{R}^{\cdots\times C}$, the last dimension is padded to a multiple of the group size $g$, reshaped into channel groups, assigned a group-wise base radius by max-absolute value or an optional \texttt{kthvalue}-based percentile on $|X|$, projected by DCRP, and quantized by signed symmetric INT4 quantize--dequantize before the original linear operator is executed on the dequantized activation.
Weight quantization is static.
Weights are quantized once when the wrapper is constructed, can be packed into INT4 for serialization, and are dequantized on the fly before the linear operation.

The convolutional path differs across the two backbones.
In the CFRN wrappers, \texttt{QuantConv2d} and \texttt{QuantConv1d} use standard symmetric activation quantization in per-tensor or per-channel mode and do not invoke the token-group DCRP routine.
In the ESCNet quantizer, the dynamic COD-TDQ activation kernel includes an explicit 4D branch that maps $X\in\mathbb{R}^{B\times C\times H\times W}$ to $X'\in\mathbb{R}^{B\times HW\times C}$, applies DSTG and DCRP along the channel dimension, and restores the original layout afterward.
Accordingly, convolutional tokenization is part of the ESCNet dynamic path, rather than a shared implementation detail across both frameworks.

Other operators, including \texttt{LayerNorm}, \texttt{GroupNorm}, \texttt{Softmax}, residual additions, non-parameter reshapes, interpolation and upsampling, and output post-processing, remain on the original floating-point path used by the evaluation codebase.
After wrapper injection, the builder can also apply an auxiliary Linear I/O correction stage with modes \texttt{none}, \texttt{bias}, and \texttt{affine}.
This stage is independent of the core DSTG+DCRP quantizer.
Unless stated otherwise, all comparisons in the paper follow the common operator scope defined in this subsection.
Baseline families are summarized in Supp. Sec.~S4.1, and the reproducibility checklist is summarized in Supp. Sec.~S4.8.

\subsection{Runtime and Storage}
\label{sec:supp_runtime}

DSTG+DCRP adds one max or \texttt{kthvalue} reduction, one standard deviation, two element-wise min projections, and one round--clamp--dequant sequence per token-group.
The resulting bookkeeping scales linearly with the number of activation elements.
For tokenized tensors, the activation-side cost is $O(BTC)$, where $B$ is the batch size, $T$ is the number of tokens, and $C$ is the channel dimension.
In the ESCNet 4D branch, the same linear scaling applies after permutation and reshape.
In the CFRN convolutional wrappers, convolutional activations follow the ordinary per-tensor or per-channel QDQ path and do not incur token-group reductions.
The method is architecture-agnostic and can be applied to Swin- and PVT-style COD backbones without modifying training.

\paragraph{Online quantization and fixed controls.}
The reported COD-TDQ path is online.
In the current implementation, clip radii are computed from the current activation tensor at each forward pass, and no iterative reconstruction or layer-wise search is performed during inference.
As a supplementary control, we also report a fixed offline-calibrated variant that pre-computes per-layer radii from a calibration subset and keeps them fixed during evaluation.

\paragraph{Simulated runtime and packed storage.}
All throughput numbers in this submission are measured under Triton INT4 path.
In the current wrappers, packed INT4 weights are dequantized on the fly, and \texttt{F.linear}, \texttt{F.conv1d}, and \texttt{F.conv2d} are executed in the selected floating-point \texttt{compute\_dtype} (\texttt{fp16} or \texttt{fp32}).
These measurements therefore characterize the present quantize--dequantize software path rather than native INT4 kernel execution.

\subsection{Default Hyperparameters}

The main paper uses a single shared setting across all datasets and both backbones, namely DSTG group size $g=32$ (Eq.~\eqref{eq:direct_sum}), DCRP resolution bound $\tau=1.0$ (Eq.~\eqref{eq:tau_constraint}), and zero-bin mass bound $\mathrm{zr}=0.2$ (Eq.~\eqref{eq:zr_constraint}).
On the dynamic token-group path, these correspond to \texttt{act\_group\_size}$=32$, \texttt{step\_over\_std}$=1.0$, and \texttt{token\_zero\_ratio}$=0.2$.
Unless a layer-specific override is explicitly supplied, the same setting is reused across datasets and both backbones.

\begin{algorithm}[t]
\caption{Forward of \texttt{QuantLinear} with DSTG + DCRP}
\label{alg:quantlinear}
\centering
{\parbox{0.95\linewidth}{
\textbf{Input:} activation $x\in\mathbb{R}^{\cdots\times C}$, bits $(w,a)$, group size $g$, and bounds $(\tau,\mathrm{zr})$.\\
\textbf{Output:} $y=\mathrm{Linear}(\hat{x},\hat{W})$ under simulated quantization.\\[0.25em]
1.\; Pad $x$ on the last dimension to $C_{\text{pad}}$ such that $C_{\text{pad}}\bmod g=0$.\\
2.\; Reshape $x_g \leftarrow \mathrm{view}(x,\,\cdots,\,C_{\text{pad}}/g,\,g)$.\\
3.\; \textbf{DSTG:} compute $c^{\text{base}} \leftarrow \|x_g\|_\infty$ or $Q_p(|x_g|)$ along the last dimension.\\
4.\; \textbf{DCRP:} if both $\tau$ and $\mathrm{zr}$ are provided, then\\
\hspace*{1.2em}(i)\; $\sigma\leftarrow \mathrm{Std}(x_g)$ with \texttt{unbiased=False}, and $c^{(\tau)}\leftarrow q_{\max}\tau\sigma$.\\
\hspace*{1.2em}(ii)\; $\mathrm{thr}\leftarrow Q_{\mathrm{zr}}(|x_g|)$ via a \texttt{kthvalue} quantile, and $c^{(\mathrm{zr})}\leftarrow 2q_{\max}\mathrm{thr}$.\\
\hspace*{1.2em}(iii)\; $c \leftarrow \min(c^{\text{base}},c^{(\tau)},c^{(\mathrm{zr})})$. Otherwise, $c\leftarrow c^{\text{base}}$.\\
5.\; Clip $\tilde{x}\leftarrow \operatorname{clip}(x_g,-c,c)$.\\
6.\; Step $\Delta\leftarrow \max(c/q_{\max},10^{-8})$.\\
7.\; Quantize $q\leftarrow \operatorname{clip}(\operatorname{round}(\tilde{x}/\Delta),\ q_{\min},q_{\max})$.\\
8.\; Dequantize $\hat{x}_g\leftarrow q\cdot \Delta$.\\
9.\; Reshape $\hat{x}_g$ back and unpad to obtain $\hat{x}$.\\
10.\; Compute $y=\mathrm{Linear}(\hat{x},\hat{W})$, where $\hat{W}$ is dequantized from static $w$-bit weights and the operator runs in floating-point \texttt{compute\_dtype}.
}}
\end{algorithm}

\section{Method Details}
\label{sec:supp_method}

\subsection{Interpretation of $\tau$}
\label{sec:clarify}

The symbol $\tau$ appears in several PTQ papers with different meanings.
In COD-TDQ, $\tau$ upper-bounds the step-to-dispersion ratio $\eta=\Delta/\sigma$ for uniform signed activation quantization.
It acts on pre-Linear activations through the clip-radius constraint in Eq.~\eqref{eq:tau_constraint}, and it enters the deterministic projection of Eq.~\eqref{eq:dcrp}.
This usage differs from $\tau$ parameters introduced for attention-map quantization or adaptive granularity in other settings, which act after Softmax and are typically selected by calibration search rather than by a fixed stability bound.

\subsection{Quantile Implementation}
\label{sec:supp_quantile}

When percentile clipping in Eq.~\eqref{eq:DSTG_clip} is enabled, the implementation uses a kthvalue-based quantile for efficiency and determinism.
For a group of size $g$ and target percentile $p\in(0,1]$, the rank is computed as $k=\lceil pg\rceil$, and kthvalue uses this 1-indexed rank directly.
The resulting quantile is therefore discrete and group-size dependent.
For example, when $g=32$ and $p=0.99$, one obtains $k=32$, which reduces to the group maximum.
Ties inherit the deterministic ordering returned by kthvalue.
The common COD-TDQ description in this supplementary uses max-absolute group radii.

\subsection{DCRP Trade-off and Bounds}
\label{sec:supp_dcrp_guarantee}

\paragraph{Clipping and rounding.}
DCRP trades rounding and zeroization error against clipping error, which is especially pronounced at 4-bit precision.
For element-wise quantization with $\tilde{x}=\operatorname{clip}(x,-c,c)$ and $\hat{x}=Q(\tilde{x},\Delta)$, the total distortion decomposes as
\[
x-\hat{x}=(x-\tilde{x})+(\tilde{x}-\hat{x}),
\]
where the first term is the clipping error and the second term is the rounding error.
In the implemented token-group path, DCRP improves W4A4 COD robustness by reducing discretization and zero-bin collapse through the bounds on $\Delta/\sigma$ and the empirical zero-bin mass.
This is achieved at the cost of controlled tail clipping when the clip radius is reduced.

\paragraph{Lemma 1 (Joint-constraint satisfaction).}
Assume that both $\tau$ and $\mathrm{zr}$ are provided for token-group $k$.
The implemented projection sets
\begin{equation}
c_k=\min\!\big(c_k^{\text{base}},\, q_{\max}\tau\sigma_k,\, 2q_{\max}Q_{\mathrm{zr}}(|x_k|)\big),
\end{equation}
where $\sigma_k=\operatorname{Std}(x_k)$ is computed with \texttt{unbiased=False}.
Then $\Delta_k\le \tau\sigma_k$.
The empirical zero-bin threshold also satisfies $\Delta_k/2 \le Q_{\mathrm{zr}}(|x_k|)$, so the empirical zero-bin mass is bounded approximately by $\mathrm{zr}$ up to discrete-rank slack from the \texttt{kthvalue} quantile.

\textbf{Proof.}
The first claim follows directly from $c_k\le q_{\max}\tau\sigma_k$ and $\Delta_k=c_k/q_{\max}$.
For the second claim, the implementation enforces $c_k\le 2q_{\max}Q_{\mathrm{zr}}(|x_k|)$, which implies $\Delta_k/2 \le Q_{\mathrm{zr}}(|x_k|)$.
Because $Q_{\mathrm{zr}}$ is a rank-based empirical quantile, the resulting zero-bin statement is discrete rather than continuum-exact.

\paragraph{Proposition 1 (Rounding-noise bound under C1).}
Let $e^{\text{rnd}}_{k}=\hat{x}_{k}-\tilde{x}_{k}$ be the rounding error in Eqs.~\eqref{eq:clipx}--\eqref{eq:dequant} for a group of size $g$.
Then $\|e^{\text{rnd}}_{k}\|_{\infty}\le \Delta_{k}/2$, and thus
\begin{equation}
\|e^{\text{rnd}}_{k}\|_2 \le \sqrt{g}\,\frac{\Delta_{k}}{2}
\le \sqrt{g}\,\frac{\tau\sigma_{k}}{2},
\label{eq:rnd_bound}
\end{equation}
whenever the projection enforces $\Delta_k\le\tau\sigma_k$.
If rounding is modeled as uniformly distributed noise, then.
\[
\mathbb{E}\|e^{\text{rnd}}_{k}\|_2^2 \le \frac{g\,\Delta_{k}^2}{12}
\le \frac{g\,\tau^2\sigma_{k}^2}{12}.
\]

\textbf{Proof.}
Each scalar rounding error is bounded by $\Delta_k/2$.
The $\ell_2$ bound follows by summing $g$ bounded components.
The expectation uses the classical $\Delta^2/12$ variance of uniform quantization noise.

\paragraph{Discussion.}
Eq.~\eqref{eq:rnd_bound} explains why the step-to-dispersion bound is COD-relevant.
COD cues are often small relative to $\sigma$, so keeping $\Delta/\sigma$ bounded limits the discretization of weak boundary signals.
The zero-bin constraint targets the zero-bin mass collapse diagnosed in the main paper.
By shrinking the clip radius when too much mass concentrates near zero, weak but structured responses are less likely to be annihilated.
In the COD failure regime studied here, controlled tail clipping can be less harmful than a coarse step size because heavy-tail outliers are frequently associated with background texture tokens or sporadic attention spikes that mainly inflate the effective range.

\subsection{Diagnostic Protocol}
\label{sec:testable_diagnosis}

The mechanism analysis in the main paper uses post hoc forward-pass measurements under the common quantization protocol.
The construction of boundary-heavy activation sets, image-to-token mapping, and layer-wise aggregation is specified in Supp. Secs.~S4.4 and~S4.5.
Ground-truth boundary sets are used only for post hoc diagnosis and are never used for calibration, range selection during inference, or prediction generation.

\paragraph{Range disparity.}
Layers with large range disparity $\mathcal{D}(X)$ in Eq.~\eqref{eq:range_disparity} show inflated step sizes and larger boundary-heavy $\rho_0$ under a shared-range quantizer.
Measuring $\mathcal{D}(X)$, $\Delta$, $\eta$, and $\rho_0$ across layers therefore provides a direct diagnostic view of W4A4 failure.

\paragraph{Token-local ranges.}
Replacing a tensor-wise range with token-local ranges, while keeping the same 4-bit grid, decreases the effective $\Delta$ for the majority of tokens and reduces boundary-heavy $\rho_0$.
This reduction is expected to track improvements in COD metrics such as \(S_\alpha\).

\paragraph{Clipping and zeroization.}
Under W4A4, decreasing the clip radius reduces $\Delta$ and therefore reduces $\rho_0$, but it also increases the clipping of extreme values.
If COD failure is driven primarily by boundary-cue annihilation, moderate additional tail clipping would be less harmful than an increase in boundary-heavy $\rho_0$.

\paragraph{Activation precision.}
Under the same quantization protocol, increasing activation precision from 4-bit to 8-bit recovers accuracy more reliably than increasing weight precision alone.
This trend is consistent with Table~\ref{tab:motivation} in the main paper and can be validated per layer by monitoring $\Delta$, $\eta$, and $\rho_0$ under the protocol of Supp. Secs.~S4.4 and~S4.5.

\section{Additional Experiments}
\label{sec:supp_exps}

\subsection{Hyperparameter Sensitivity}
\label{sec:exp_hyper}

\paragraph{Protocol.}
We study the sensitivity of the three key hyperparameters in COD-TDQ, namely DSTG group size $g$ (Eq.~\eqref{eq:direct_sum}), DCRP resolution bound $\tau$ (Eq.~\eqref{eq:tau_constraint}), and DCRP zero-bin mass bound $\mathrm{zr}$ (Eq.~\eqref{eq:zr_constraint}).

\paragraph{Shared setting across datasets and backbones.}
The main paper reports the shared setting $(g,\tau,\mathrm{zr})=(32,1.0,0.2)$ across all datasets and both CFRN and ESCNet.

We add a label-free self-selection procedure without retraining. For each candidate \((g,\tau,\mathrm{zr})\), we run a calibration forward pass and compute a score that combines intermediate feature drift against FP32 features, the step-to-dispersion ratio, the zero-bin mass, and the zero-bin mass estimated from FP32 predictions. The selected configuration is the one with the lowest score, which matches the setting used in the paper.

\subsection{Checkpoint Storage Footprint}
\label{sec:exp_size}

\begin{table}[t]
\centering
\caption{\textbf{Checkpoint storage footprint (\texttt{.pth}).}}
\label{tab:model_size}
\setlength{\tabcolsep}{6pt}
\renewcommand{\arraystretch}{1.15}
\begin{tabular}{lccc}
\toprule
Model & FP32 (MB) & Packed INT4 weights (MB) & Ratio \\
\midrule
CFRN~\cite{song2025cfrn} & 755.4 & 110.0 & 6.9$\times$ \\
ESCNet~\cite{ye2025escnet} & 381.8 & 101.5 & 3.8$\times$ \\
\bottomrule
\end{tabular}
\end{table}

While the main paper focuses on accuracy preservation under W4A4 simulated quantization, packing 4-bit weights also reduces serialized checkpoint size.
The numbers in \cref{tab:model_size} are storage references only.
They do not include activation-side buffers or dynamic per-group statistics in the current software path, and they should not be interpreted as evidence of native INT4 kernel execution.
The deployment-facing distinction between storage, simulated runtime, and native INT4 execution is summarized in Supp. Sec.~S4.7.

\subsection{Full Experimental Tables}
\label{sec:full_data}

Supplementary experimental tables are provided in \cref{tab:escnet_w4a4_full}.
Table~\ref{tab:escnet_w4a4_full} reports ESCNet W4A4 results on all four datasets under the common operator scope of Supp. Sec.~S1.1.

For the CFRN main comparison across all PTQ baselines, see Table~\ref{tab:cfrn_w4a4} in the main paper.

\begin{table*}[t]
\centering
\caption{\textbf{W4A4 post-training quantization results on ESCNet across four COD benchmarks.}}
\label{tab:escnet_w4a4_full}
\scriptsize
\setlength{\tabcolsep}{2.8pt}
\renewcommand{\arraystretch}{1.12}
\resizebox{\linewidth}{!}{
\begin{tabular}{l | ccccc | ccccc | ccccc | ccccc}
\toprule
\multirow{2}{*}{Method}
& \multicolumn{5}{c|}{\textbf{CAMO}}
& \multicolumn{5}{c|}{\textbf{CHAMELEON}}
& \multicolumn{5}{c|}{\textbf{COD10K}}
& \multicolumn{5}{c}{\textbf{NC4K}} \\
\cmidrule(lr){2-6}
\cmidrule(lr){7-11}
\cmidrule(lr){12-16}
\cmidrule(lr){17-21}
& $S_\alpha$ & $F^{\omega}_{\beta}$ & $E_m$ & $F^m_\beta$ & MAE
& $S_\alpha$ & $F^{\omega}_{\beta}$ & $E_m$ & $F^m_\beta$ & MAE
& $S_\alpha$ & $F^{\omega}_{\beta}$ & $E_m$ & $F^m_\beta$ & MAE
& $S_\alpha$ & $F^{\omega}_{\beta}$ & $E_m$ & $F^m_\beta$ & MAE \\
\midrule
FP32
& .8755 & .8488 & .9372 & .8742 & .0408
& .8985 & .8672 & .9525 & .8850 & .0226
& .8734 & .8082 & .9421 & .8369 & .0204
& .8929 & .8640 & .9451 & .8869 & .0278 \\
Naive W8A8
& .8755 & .8488 & .9372 & .8742 & .0408
& .8988 & .8673 & .9517 & .8852 & .0226
& .8734 & .8082 & .9421 & .8369 & .0204
& .8927 & .8637 & .9448 & .8867 & .0279 \\
Naive W4A8
& .8669 & .8386 & .9296 & .8670 & .0445
& .8930 & .8559 & .9425 & .8744 & .0227
& .8655 & .7978 & .9351 & .8290 & .0219
& .8881 & .8581 & .9411 & .8834 & .0292 \\
Naive W4A4
& .4680 & .1802 & .3998 & .2514 & .1687
& .4505 & .0815 & .3441 & .1709 & .1400
& .5455 & .2500 & .5015 & .3381 & .0808
& .5762 & .3675 & .5618 & .4590 & .1201 \\
\midrule
ORQ-ViT
& .4682 & .1805 & .4001 & .2516 & .1689
& .4544 & .0885 & .3370 & .1710 & .1368
& .5456 & .2502 & .5017 & .3383 & .0811
& .5764 & .3677 & .5619 & .4593 & .1203 \\
post-GELU
& .4716 & .1836 & .4035 & .2544 & .1681
& .4546 & .0817 & .3281 & .1711 & .1370
& .5478 & .2530 & .5047 & .3418 & .0805
& .5808 & .3745 & .5663 & .4670 & .1187 \\
PQ-SAM
& .4924 & .2342 & .4429 & .3153 & .1613
& .4774 & .1403 & .3850 & .1987 & .1320
& .5700 & .3053 & .5436 & .4067 & .0758
& .6085 & .4287 & .6072 & .5255 & .1107 \\
NoisyQuant
& .6511 & .5414 & .7066 & .6298 & .1378
& .6037 & .4950 & .7108 & .6239 & .1012
& .6618 & .5215 & .7055 & .5981 & .0695
& .7143 & .6513 & .8034 & .6843 & .0934 \\
PTQ4ViT
& .6748 & .5676 & .7234 & .6312 & .1219
& .6257 & .5371 & .7036 & .6317 & .0969
& .6903 & .5221 & .7186 & .6022 & .0666
& .7232 & .6577 & .8179 & .6995 & .0795 \\
FIMA-Q
& .6859 & .5784 & .7491 & .6443 & .1209
& .6947 & .5668 & .7120 & .6422 & .0767
& .7099 & .5437 & .7211 & .6175 & .0591
& .7477 & .6617 & .8211 & .7101 & .0776 \\
PTQ4SAM
& .7209 & .6516 & .7984 & .7016 & .0988
& .6550 & .6108 & .6952 & .7685 & .0956
& .7257 & .5987 & .8104 & .6456 & .0523
& .7683 & .7012 & .8433 & .7413 & .0616 \\
IGQ-ViT
& .7511 & .6925 & .8393 & .7499 & .0827
& .7723 & .7076 & .8635 & .7589 & .0495
& .7533 & .6429 & .8573 & .6897 & .0454
& .8182 & .7268 & .8777 & .7901 & .0547 \\
RepQ-ViT
& .7707 & .7014 & .8482 & .7514 & .0786
& .8006 & .7218 & .8737 & .7715 & .0467
& .7755 & .6485 & .8602 & .6954 & .0422
& .8182 & .7510 & .8831 & .7911 & .0515 \\
\textbf{Ours}
& \textbf{.8528} & \textbf{.8197} & \textbf{.9158} & \textbf{.8517} & \textbf{.0502}
& \textbf{.8899} & \textbf{.8517} & \textbf{.9557} & \textbf{.8706} & \textbf{.0234}
& \textbf{.8539} & \textbf{.7786} & \textbf{.9252} & \textbf{.8120} & \textbf{.0245}
& \textbf{.8813} & \textbf{.8485} & \textbf{.9358} & \textbf{.8757} & \textbf{.0312} \\
\bottomrule
\end{tabular}
}
\end{table*}

We next provide layer-wise parameter diagnostics and additional qualitative comparisons that complement these aggregate tables.

\subsection{Layer-wise Parameter Statistics and Additional Qualitative Comparisons}
\label{sec:supp_weight_diagnostics}

\begin{figure*}[t]
\centering
\includegraphics[width=0.97\textwidth]{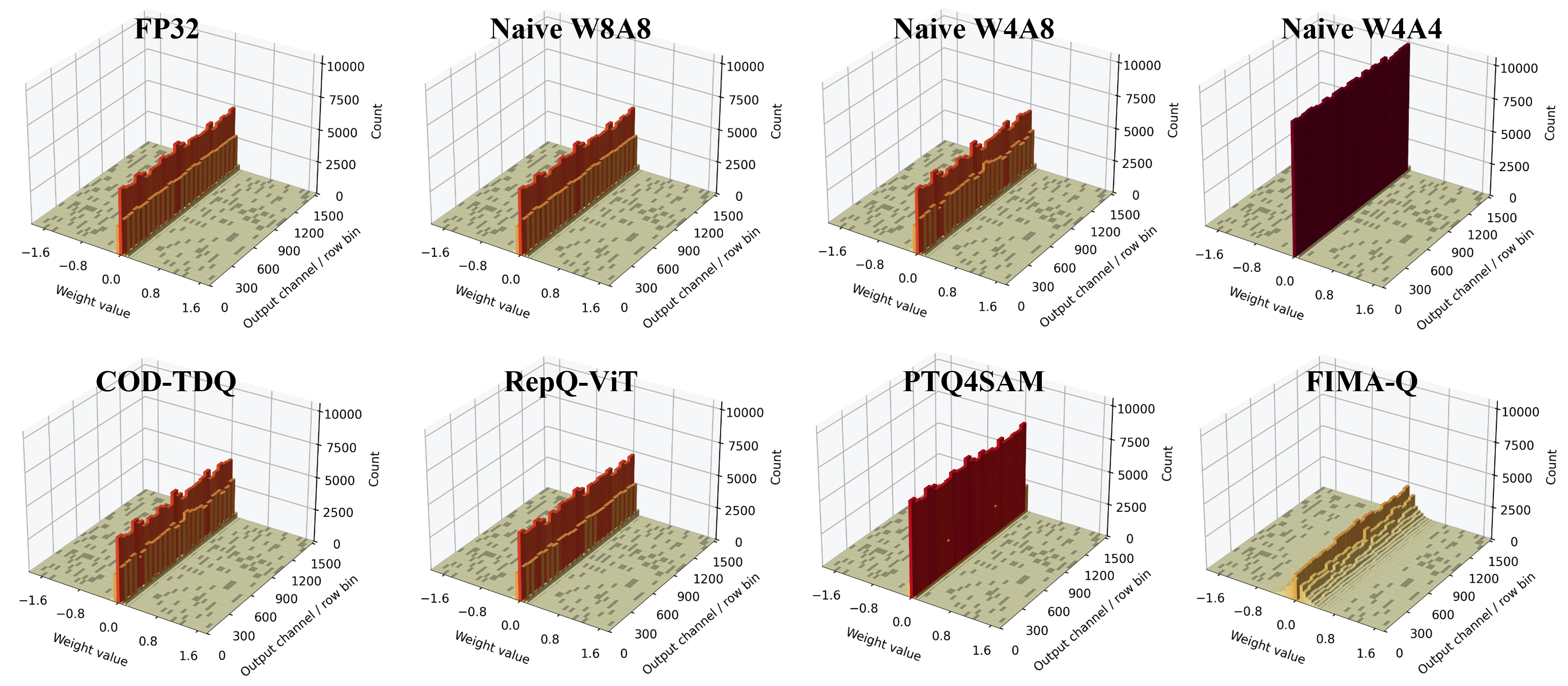}
\caption{\textbf{Layer-wise parameter histogram for \texttt{swin.layers2.blocks10.qkv}.} The three axes denote weight value, output channel, and count. Each panel corresponds to one model, ordered as FP32, Naive W8A8, Naive W4A8, Naive W4A4, Ours, RepQ-ViT, PTQ4SAM, and FIMA-Q. This visualization complements the activation-side analysis by showing how the parameter distribution of a sensitive qkv projection changes across quantization methods.}
\label{fig:supp_qkv_hist}
\end{figure*}

\begin{figure*}[t]
\centering
\includegraphics[width=0.97\textwidth]{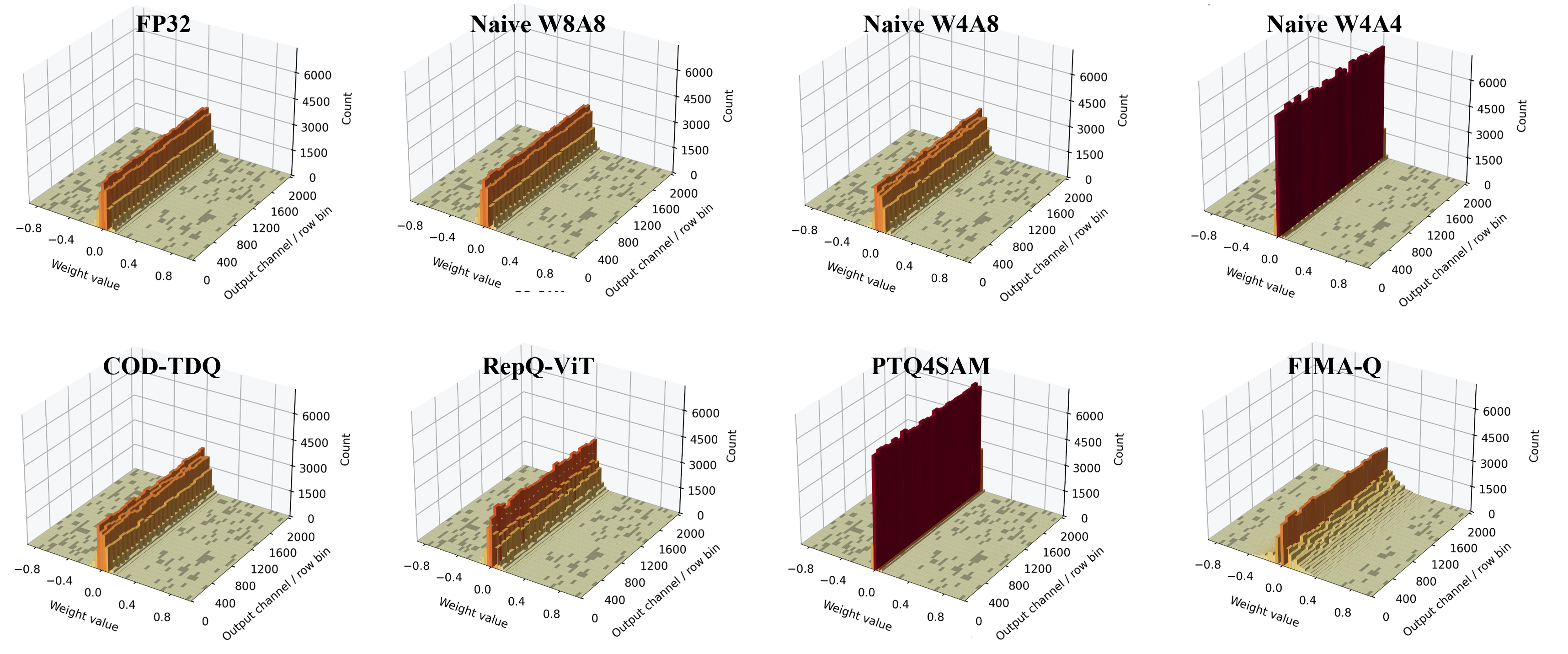}
\caption{\textbf{Layer-wise parameter histogram for \texttt{swin.layers2.blocks0.mlp1}.} The three axes denote weight value, output channel, and count. Each panel corresponds to one model, ordered as FP32, Naive W8A8, Naive W4A8, Naive W4A4, Ours, RepQ-ViT, PTQ4SAM, and FIMA-Q. Together with \cref{fig:supp_qkv_hist}, this figure shows how the parameter distribution behaves at a representative MLP projection across quantization methods.}
\label{fig:supp_fc1_hist}
\end{figure*}

To complement the activation-side diagnostics in the main paper, we provide layer-wise parameter histograms for two representative operators, namely \texttt\texttt{swin.layers2.blocks10.qkv} and \texttt{swin.blocks0.mlp1}. Figures~\ref{fig:supp_qkv_hist} and~\ref{fig:supp_fc1_hist} visualize the parameter distributions across FP32, Naive W8A8, Naive W4A8, Naive W4A4, Ours, RepQ-ViT, PTQ4SAM, and FIMA-Q. These views offer a compact cross-model diagnostic at two sensitive layers and complement the activation-oriented mechanism analysis in the main paper.

Tables~\ref{tab:supp_weight_stats_qkv} and~\ref{tab:supp_weight_stats_mlp} report the corresponding numerical summaries, including the mean, standard deviation, 99th percentile, and absolute maximum of the exported parameter records. 

\begin{figure*}[t]
\centering
\includegraphics[width=0.98\textwidth]{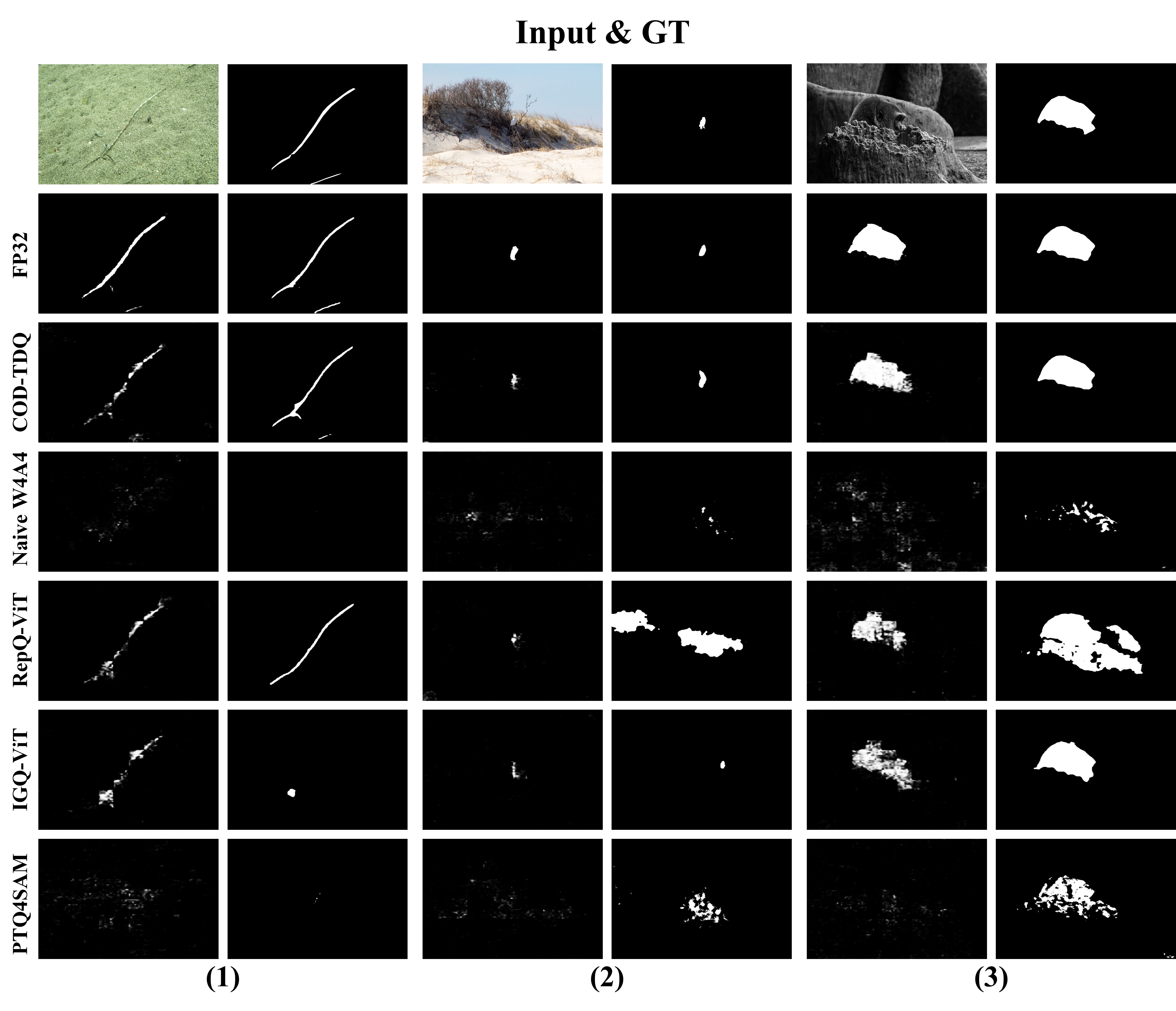}
\caption{\textbf{Additional qualitative output comparison on representative COD examples.} Each example shows the input image, the ground-truth mask, and the predictions produced by FP32, naive W4A4, representative PTQ baselines, and COD-TDQ under the same evaluation protocol. This figure complements Fig.~\ref{fig:qualitative} in the main paper by providing additional cases beyond the main-page qualitative comparison. The figure is divided into groups (1), (2), and (3), with CFRN results on the left and ESCNet results on the right.}
\label{fig:supp_output_compare}
\end{figure*}

Figure~\ref{fig:supp_output_compare} further extends the qualitative evaluation with additional representative examples under the same W4A4 protocol. This comparison should be read together with Fig.~\ref{fig:qualitative} in the main paper and provides an output-level counterpart to the layer-wise histogram diagnostics.

\begin{table*}[t]
\centering
\caption{\textbf{Parameter summary statistics for the qkv-related export block (layer \texttt{blocks.10.attn.qkv}).} }
\label{tab:supp_weight_stats_qkv}
\small
\setlength{\tabcolsep}{4.2pt}
\renewcommand{\arraystretch}{1.14}
\begin{tabular*}{\textwidth}{@{\extracolsep{\fill}} l r c c c @{}}
\toprule
Model & Mean & Std. & P99 & AbsMax \\
\midrule
FP32 & \(-6.450\times 10^{-5}\) & 0.05019 & 0.11918 & 0.45400 \\
Naive W8A8 & \(-6.950\times 10^{-5}\) & 0.05049 & 0.12203 & 0.45302 \\
Naive W4A8 & \(-7.560\times 10^{-5}\) & 0.05074 & 0.12127 & 0.45306 \\
Naive W4A4 & \(-2.860\times 10^{-4}\) & 0.18971 & 0.45260 & 1.89392 \\
Ours & \(-7.800\times 10^{-5}\) & 0.05074 & 0.12055 & 0.45400 \\
RepQ-ViT & \(-7.930\times 10^{-6}\) & 0.05055 & 0.11903 & 0.27057 \\
IGQ-ViT & \(-4.980\times 10^{-5}\) & 0.03136 & 0.07461 & 0.35816 \\
PTQ4SAM & \(-4.790\times 10^{-5}\) & 0.05261 & 0.12661 & 1.13471 \\
FIMA-Q & \(-1.480\times 10^{-5}\) & 0.01417 & 0.03398 & 0.19658 \\
\bottomrule
\end{tabular*}
\end{table*}

\begin{table*}[t]
\centering
\caption{\textbf{Parameter summary statistics for the MLP-related export block (layer \texttt{blocks.0.mlp.fc1}).} The common prefix \texttt{swin.swin\_encoder.layers.2.} is omitted from the sublayer name for readability.}
\label{tab:supp_weight_stats_mlp}
\small
\setlength{\tabcolsep}{4.2pt}
\renewcommand{\arraystretch}{1.14}
\begin{tabular*}{\textwidth}{@{\extracolsep{\fill}} l r c c c @{}}
\toprule
Model & Mean & Std. & P99 & AbsMax \\
\midrule
FP32 & \(7.620\times 10^{-4}\) & 0.04994 & 0.12023 & 0.34726 \\
Naive W8A8 & \(7.720\times 10^{-4}\) & 0.04993 & 0.12022 & 0.34732 \\
Naive W4A8 & \(7.820\times 10^{-4}\) & 0.05046 & 0.12131 & 0.34721 \\
Naive W4A4 & \(2.164\times 10^{-3}\) & 0.14529 & 0.35743 & 1.02400 \\
Ours & \(7.821\times 10^{-4}\) & 0.05046 & 0.12137 & 0.34726 \\
RepQ-ViT & \(8.130\times 10^{-5}\) & 0.05030 & 0.11789 & 0.26432 \\
IGQ-ViT & \(3.173\times 10^{-4}\) & 0.02489 & 0.05915 & 0.94216 \\
PTQ4SAM & \(3.131\times 10^{-4}\) & 0.01896 & 0.04586 & 0.13673 \\
FIMA-Q & \(1.085\times 10^{-3}\) & 0.05326 & 0.13600 & 0.58206 \\
\bottomrule
\end{tabular*}
\end{table*}

\section{Reproducibility and Scope}
\label{sec:supp_protocol}

\subsection{Baselines and Operator Scope}
\label{sec:supp_baseline_matrix}

Unless stated otherwise, all baseline numbers in Tables~\ref{tab:cfrn_w4a4}, \ref{tab:escnet_nc4k}, and \ref{tab:escnet_w4a4_full} are reported under the common operator scope defined in Supp. Sec.~S1.1.
This convention standardizes side-by-side comparison across CFRN and ESCNet.
It does not attempt to replicate every low-level implementation detail of the original baseline releases.

\begin{table}[t]
\centering
\caption{\textbf{Baseline families under the common operator scope.}}
\label{tab:supp_baseline_matrix}
\small
\setlength{\tabcolsep}{4pt}
\renewcommand{\arraystretch}{1.16}
\begin{tabularx}{\linewidth}{@{}>{\raggedright\arraybackslash}p{0.26\linewidth}X@{}}
\toprule
Family & Methods \\
\midrule
ViT PTQ baselines & NoisyQuant, PTQ4ViT, ORQ-ViT, post-GELU, FIMA-Q, IGQ-ViT, and RepQ-ViT \\
SAM-oriented PTQ transfers & PQ-SAM, PTQ4SAM, and AHCPTQ \\
Cross-domain PTQ transfers & SmoothQuant, QuaRTZ, and ARCQuant \\
\bottomrule
\end{tabularx}
\end{table}

\subsection{Variant Definitions}
\label{sec:supp_ablation_taxonomy}

Table~\ref{tab:supp_ablation_taxonomy} fixes the variant names used in the main paper and the supplementary.
In the current implementation, the default DCRP helper activates the joint $\tau+\mathrm{zr}$ projection.
Accordingly, the one-sided C1-only and C2-only rows correspond to dedicated ablation variants.

\begin{table*}[t]
\centering
\caption{\textbf{Variant definitions used in the ablation study.}}
\label{tab:supp_ablation_taxonomy}
\scriptsize
\setlength{\tabcolsep}{3.8pt}
\renewcommand{\arraystretch}{1.14}
\begin{tabularx}{\textwidth}{@{}>{\raggedright\arraybackslash}p{0.22\textwidth}>{\centering\arraybackslash}p{0.06\textwidth}>{\centering\arraybackslash}p{0.06\textwidth}>{\centering\arraybackslash}p{0.11\textwidth}>{\raggedright\arraybackslash}p{0.16\textwidth}X@{}}
\toprule
Variant & DSTG & DCRP & Active constraints & Scope & Description \\
\midrule
Naive W4A4 / Naive W4A4 (per-tensor) & No & No & -- & Shared activation range & Baseline shared-range control used in the main comparison tables \\
Per-tensor & No & No & -- & Global-scale activation QDQ & Explicit global-range ablation used in the CFRN ablation table \\
DSTG only / DSTG only (w/o DCRP) & Yes & No & -- & Common operator scope & Token-group base radii without DCRP projection \\
DCRP-Attn & No & Yes & C1 + C2 & Attention projections only & CFRN control reported in the main paper \\
DCRP only & No & Yes & C1 + C2 & All quantized operators & ESCNet control reported in the main paper \\
DCRP only (fixed offline calibration) & No & Yes & C1 + C2 & Fixed per-layer control & Supplementary control with radii pre-computed from the calibration subset \\
DCRP (C1-only) & No & Yes & C1 only & Common operator scope & One-sided stability ablation \\
DCRP (C2-only) & No & Yes & C2 only & Common operator scope & One-sided zero-bin ablation \\
COD-TDQ / Ours / DSTG + DCRP & Yes & Yes & C1 + C2 & Full common operator scope & Full method reported in the main paper and the supplementary \\
\bottomrule
\end{tabularx}
\end{table*}

\subsection{Backbone-Wise Gains}
\label{sec:supp_gain_clarification}

The strongest baseline gap is backbone-dependent.
Table~\ref{tab:supp_gain_clarification} reports the exact $S_\alpha$ difference between COD-TDQ and the strongest non-ours baseline under the same common protocol.
On CFRN, the gains range from $+.119$ to $+.161$.
On ESCNet, the corresponding gains range from $+.0631$ to $+.0893$.

\begin{table*}[t]
\centering
\caption{\textbf{Backbone-wise gain over the strongest non-ours baseline in $S_\alpha$.}}
\label{tab:supp_gain_clarification}
\small
\renewcommand{\arraystretch}{1.15}
\begin{tabular}{llccc}
\toprule
Backbone & Dataset & Strongest baseline ($S_\alpha$) & COD-TDQ ($S_\alpha$) & Gap \\
\midrule
CFRN & CAMO & RepQ-ViT (.676) & .813 & +.137 \\
CFRN & CHAMELEON & RepQ-ViT (.701) & .862 & +.161 \\
CFRN & COD10K & RepQ-ViT (.676) & .802 & +.126 \\
CFRN & NC4K & RepQ-ViT (.718) & .837 & +.119 \\
\midrule
ESCNet & CAMO & RepQ-ViT (.7707) & .8528 & +.0821 \\
ESCNet & CHAMELEON & RepQ-ViT (.8006) & .8899 & +.0893 \\
ESCNet & COD10K & RepQ-ViT (.7755) & .8539 & +.0784 \\
ESCNet & NC4K & IGQ-ViT / RepQ-ViT tie (.8182) & .8813 & +.0631 \\
\bottomrule
\end{tabular}
\end{table*}

\subsection{Boundary-Heavy Protocol}
\label{sec:supp_boundary_protocol}

To make the mechanism analysis reproducible, we define the boundary-heavy diagnostic set at the token level and then inherit that label to token-groups.
Let $M\in\{0,1\}^{H\times W}$ be the binary ground-truth mask for an image.
For analysis, define an image-space boundary band
\begin{equation}
B_r = \mathrm{Dilate}(M,r_{\text{out}})\setminus \mathrm{Erode}(M,r_{\text{in}}),
\label{eq:supp_boundary_band}
\end{equation}
where $r_{\text{in}},r_{\text{out}}\ge 0$ are the inner and outer radii of the band.
At layer $\ell$ with token grid $\Omega_\ell$, let $P_\ell(t)$ denote the image-pixel support mapped to token $t\in\Omega_\ell$.
We define the boundary occupancy of token $t$ as
\begin{equation}
\pi_\ell(t)=\frac{1}{|P_\ell(t)|}\sum_{u\in P_\ell(t)} B_r(u).
\label{eq:supp_boundary_occupancy}
\end{equation}
A token is called \emph{boundary-heavy} if $\pi_\ell(t)\ge \gamma_{\text{bdry}}$.
A token is called \emph{non-boundary} if $\pi_\ell(t)\le \gamma_{\text{nonbdry}}$.
Tokens in the interval $(\gamma_{\text{nonbdry}},\gamma_{\text{bdry}})$ are excluded from the binary comparison.
For grouped activations, each token group inherits the label of its parent token.
The boundary-heavy activation set $\mathcal{A}^{\text{bdry}}_\ell$ is formed by collecting the activations or token-group statistics associated with those boundary-heavy tokens.

Given the recorded pre-quantization activation tensor \(X_\ell\), group-wise clip radii \(c_{\ell,k}\), and derived quantities \(\Delta_{\ell,k}\), \(\sigma_{\ell,k}\), and \(\rho_{0,\ell,k}\), we compute the diagnostics described above.
These measurements are used only for analysis after inference.
They are never used by the quantizer, by calibration, or by the prediction pipeline.

\subsection{Layer-Wise Diagnostics}
\label{sec:supp_layerwise_diag}

Layer-wise diagnostics are obtained by attaching forward hooks to each quantized operator in the common scope.
For each input and operator $\ell$, the diagnostic export records (i) the pre-quantization activation tensor $X_\ell$, (ii) the projected group radii $c_{\ell,k}$, (iii) the step sizes $\Delta_{\ell,k}$, (iv) the within-group dispersions $\sigma_{\ell,k}$, and (v) the dequantized activation $\hat{X}_\ell$.
These exports support operator-wise computation of $\mathcal{D}(X_\ell)$, boundary-heavy and non-boundary summaries of $\eta_{\ell,k}=\Delta_{\ell,k}/\sigma_{\ell,k}$ and $\rho_{0,\ell,k}$, and optional clipping-rate summaries.

In the CFRN framework, \texttt{CalibrationContext} already collects Linear input statistics and Linear I/O pairs.
Group radii, step sizes, zero-bin mass, and clipping rate are exported from the quant wrappers during the diagnostic pass.
Convolutional diagnostics in CFRN require dedicated Conv hooks.

\begin{table}[t]
\centering
\caption{\textbf{Layer-wise quantities used by the diagnostic export.}}
\label{tab:supp_layerwise_diag}
\small
\setlength{\tabcolsep}{4pt}
\renewcommand{\arraystretch}{1.16}
\begin{tabularx}{\linewidth}{@{}>{\raggedright\arraybackslash}p{0.29\linewidth}>{\centering\arraybackslash}p{0.18\linewidth}X@{}}
\toprule
Diagnostic item & Directly recoverable from the current code & Note \\
\midrule
Pre-quant Linear input \(X_\ell\) & Partially & Linear inputs are visible to the calibration hook, but full-tensor archival uses an explicit export path \\
Per-channel \texttt{absmax}, \texttt{mean\_abs}, and \texttt{rms} & Yes & Collected for Linear inputs when per-channel activation statistics are enabled \\
Linear input-output sample pairs \((x,y)\) & Yes & At most 256 sampled rows are cached per Linear module by default \\
Projected group radii \(c_{\ell,k}\) and step sizes \(\Delta_{\ell,k}\) & No & Exported from the quant wrapper during the diagnostic forward pass \\
Zero-bin mass \(\rho_{0,\ell,k}\) and clipping rate & No & Derived from wrapper-side exports of group statistics \\
Convolutional diagnostics & No, in the CFRN helper & Dedicated Conv hooks are required when convolutional operators are analyzed \\
\bottomrule
\end{tabularx}
\end{table}

\subsection{Calibration Robustness}
\label{sec:supp_calib_robustness}

For methods that use offline calibration, the paper reports results for one fixed 128-image subset.
Additional sweeps over calibration subset size or random seed are outside the scope of the present tables.
COD-TDQ itself does not require activation calibration during inference.

\subsection{Runtime and Storage Scope}
\label{sec:supp_scope_accounting}

Deployment-facing claims involve four separate axes, namely simulated W4A4 evaluation, packed checkpoint size, runtime under the current software path, and native INT4 kernel execution.
The present submission establishes the first three.
Table~\ref{tab:supp_scope_accounting} summarizes this distinction.

\begin{table}[t]
\centering
\caption{\textbf{Runtime and storage scope of the present evidence.}}
\label{tab:supp_scope_accounting}
\small
\setlength{\tabcolsep}{4pt}
\renewcommand{\arraystretch}{1.16}
\begin{tabularx}{\linewidth}{@{}>{\raggedright\arraybackslash}p{0.29\linewidth}>{\centering\arraybackslash}p{0.10\linewidth}X@{}}
\toprule
Evidence item & Present & What it supports \\
\midrule
Simulated W4A4 evaluation under the current software path & Yes & Accuracy numbers and software-path runtime under the quantize--dequantize evaluation pipeline \\
Packed INT4 checkpoint size & Yes & Storage reduction for serialized weights only, see Supp. Sec.~S3.2 \\
Packed INT4 with on-the-fly floating-point compute & Yes & The wrappers dequantize packed weights and execute \texttt{linear} and \texttt{conv} in \texttt{fp16} or \texttt{fp32}, which is not native INT4 execution \\
Online DSTG and DCRP reductions in the current path & Yes & Algorithmic overhead from max, \texttt{kthvalue}, std, and min operations on activation groups \\
Native INT4 kernel execution & No & Outside the evidence provided by the present tables \\
\bottomrule
\end{tabularx}
\end{table}

\subsection{Reproducibility Checklist}
\label{sec:supp_exact_settings}

Table~\ref{tab:supp_operator_inventory} complements the common evaluation protocol with an operator-family inventory. 
To facilitate exact reproducibility, the following implementation details and experimental settings should be fixed or explicitly disclosed.

\paragraph{Base-radius instantiation for Eq.~\eqref{eq:DSTG_clip}.}
The common COD-TDQ description used in this submission adopts the maximum absolute value within each group as the base radius. 
The implementation additionally provides an optional alternative based on a \texttt{kthvalue}-based percentile computed on the absolute activation magnitude \(|x|\).

\paragraph{Numerical conventions.}
Quantization follows a signed symmetric formulation with \(q_{\min}=-2^{b-1}\) and \(q_{\max}=2^{b-1}-1\).
The quantization scale is computed as \(\max(c/q_{\max}, 10^{-8})\).
Group standard deviation is calculated with \texttt{unbiased=False} and an additive stabilization term of \(10^{-12}\).
Rounding is performed using \texttt{torch.round}, and all clipping values are clamped to a minimum of \(10^{-8}\).

\paragraph{Calibration and I/O statistics collection.}
During calibration in the CFRN helper, per-channel statistics including \texttt{absmax}, \texttt{mean\_abs}, and \texttt{rms} are collected for Linear-layer inputs only.
Optional row subsampling may limit the collected samples to at most 4096 rows.
For Linear I/O correction, the system caches up to 256 sampled rows per module.
By default, convolutional-layer I/O statistics are not collected.

\paragraph{Evaluation preprocessing and command-line protocol.}
The reference CFRN evaluation uses image resizing to 384 pixels together with ImageNet normalization.
The reference ESCNet evaluation uses resizing to 416 pixels with the same ImageNet normalization.
Prediction generation and metric computation follow the original public evaluation pipelines released with the respective backbone repositories.

\paragraph{Reference environments and default seeds.}
The backbone repositories rely on different reference environments.
CFRN uses Python~3.7, PyTorch~1.6.0, torchvision~0.7.0, and CUDA~10.2.
ESCNet uses Python~3.11 with CUDA~\(\ge\)12.4.
The public default seeds are \texttt{SEED=0} for CFRN and \texttt{rand\_seed: 42} for ESCNet.

\paragraph{Pattern-based overrides and auxiliary hooks.}
The quantization builder supports pattern-based configuration options, including 
\texttt{skip\_patterns}, \texttt{w8\_patterns}, and \texttt{keep\_a8\_patterns}.
It also allows module-name pattern rules for weight group size, weight clipping percentile, and activation group size.
An optional per-layer JSON override stage can further refine the configuration.
In addition, an auxiliary Linear I/O correction hook is provided with three selectable modes: \texttt{none}, \texttt{bias}, and \texttt{affine}.

\begin{table}[t]
\centering
\caption{\textbf{Common operator families under the shared evaluation protocol.}}
\label{tab:supp_operator_inventory}
\small
\setlength{\tabcolsep}{4pt}
\renewcommand{\arraystretch}{1.16}
\begin{tabularx}{\linewidth}{@{}>{\raggedright\arraybackslash}p{0.39\linewidth}>{\centering\arraybackslash}p{0.12\linewidth}X@{}}
\toprule
Operator family & Quantized & Common rule \\
\midrule
Attention \texttt{Linear} projections, including q, k, v, and output projections when present & Yes & Quantized under W4A4 with the COD-TDQ or baseline-specific common-protocol path \\
MLP and feed-forward \texttt{Linear} projections & Yes & Quantized under the common operator scope \\
\texttt{Conv2d} and \texttt{Conv1d} projections or decoder convolutional layers when present & Yes & Included in the common operator scope. In CFRN, they use per-tensor or per-channel activation QDQ, while the ESCNet dynamic path additionally supports 4D token-group activation QDQ \\
\texttt{LayerNorm} and \texttt{GroupNorm} & No & Kept on the floating-point path. They can be explicitly wrapped to FP32 when the corresponding safety flags are enabled \\
\texttt{Softmax} & No & Kept on the original floating-point path in the common protocol \\
Element-wise residuals, reshapes, interpolation and upsampling, output post-processing & No & Left unchanged relative to the original evaluation pipeline \\
\bottomrule
\end{tabularx}
\end{table}

\end{document}